\pdfoutput=1

\documentclass[11pt]{article}

\usepackage[preprint]{acl}

\usepackage{times}
\usepackage{latexsym}

\usepackage[T1]{fontenc}

\usepackage[utf8]{inputenc}

\usepackage{microtype}

\usepackage{inconsolata}

\usepackage{graphicx}

\usepackage{booktabs}
\usepackage{color}
\usepackage{cuted}
\usepackage{capt-of}
\usepackage{tablefootnote}
\usepackage{multirow} 
\usepackage{footnote}
\usepackage{enumitem}
\usepackage{adjustbox}
\usepackage{subcaption}
\usepackage{caption}
\usepackage{pifont}
\usepackage{algorithm}
\usepackage{xcolor}
\usepackage{wrapfig}
\usepackage{sidecap}
\usepackage{bbm}
\usepackage{tabularx}
\usepackage{amsmath}
\usepackage{amssymb}
\usepackage{multicol}
\usepackage{colortbl}
\usepackage{makecell}
\usepackage{algpseudocode}
\usepackage{arydshln}

\newcommand{\ie}{\textit{i}.\textit{e}.}
\newcommand{\eg}{\textit{e}.\textit{g}.} 

\definecolor{lightgray}{RGB}{240,240,240}

\title{Cross-Modal Masked Compositional Concept Modeling for Enhancing Visio-Linguistic Compositionality}

\author{
 \textbf{Wei Li\textsuperscript{1}},
 \textbf{Zhen Huang\textsuperscript{2}},
 \textbf{Xinmei Tian\textsuperscript{1$\dag$}}
 \\
 \textsuperscript{1}MoE Key Laboratory of Brain-inspired \\ Intelligent Perception and Cognition, University of Science and Technology of China
 \\
 \textsuperscript{2}Independent Researcher
 \\
 \texttt{lwzkd@mail.ustc.edu.cn},
 \texttt{xinmei@ustc.edu.cn}
}

\begin{document}
\maketitle

\renewcommand{\thefootnote}{\fnsymbol{footnote}}
\footnotetext[2]{Corresponding author.}
\renewcommand{\thefootnote}{\arabic{footnote}} 

\begin{abstract}
    Contrastively trained vision-language models like CLIP, have made remarkable progress in learning joint image-text representations, but still face challenges in compositional understanding. They often exhibit a ``bag-of-words'' behavior—struggling to capture the object relations, attribute-object bindings, and word order dependencies. This limitation arises not only from the reliance on global, single-vector representations for optimization, but also from the insufficient exploitation and modeling of the rich compositional information inherently present in paired image text data. In this work, we propose \textbf{MACCO} (\textbf{MA}sked \textbf{C}ompositional \textbf{C}oncept M\textbf{O}deling), a framework that masks compositional concepts in one modality and reconstructs them conditioned on the full contextual information from the other, enabling the model to capture and align cross-modal compositional structures more effectively. To facilitate this process, we introduce two auxiliary objectives that jointly align and regularize masked features both inter-modally and intra-modally. Extensive experiments on five compositional benchmarks, along with in-depth analyses, demonstrate that our approach not only significantly enhances compositionality in VLMs but also improves their ability to capture syntactic structure and linguistic information. Additionally, the improved compositionality also benefits text-to-image generation and multimodal large language model. Code is available at https://github.com/hiker-lw/MACCO.

\end{abstract}

\section{Introduction}
\label{sec:intro}
\vspace{-1mm}
Vision-language foundation models like CLIP \cite{radford2021learning} have significantly advanced multimodal learning by aligning images and texts in a shared semantic space via contrastive learning, and have been widely adopted in tasks such as image-text retrieval \cite{koukounas2024jina, chen2023altclip}, VQA \cite{zhu2023minigpt,liu2023visual}, video understanding \cite{wasim2023vita}, and text-to-image generation \cite{ramesh2022hierarchical}.

\begin{figure}[t]
    \centering
    \includegraphics[width=1.0\linewidth]{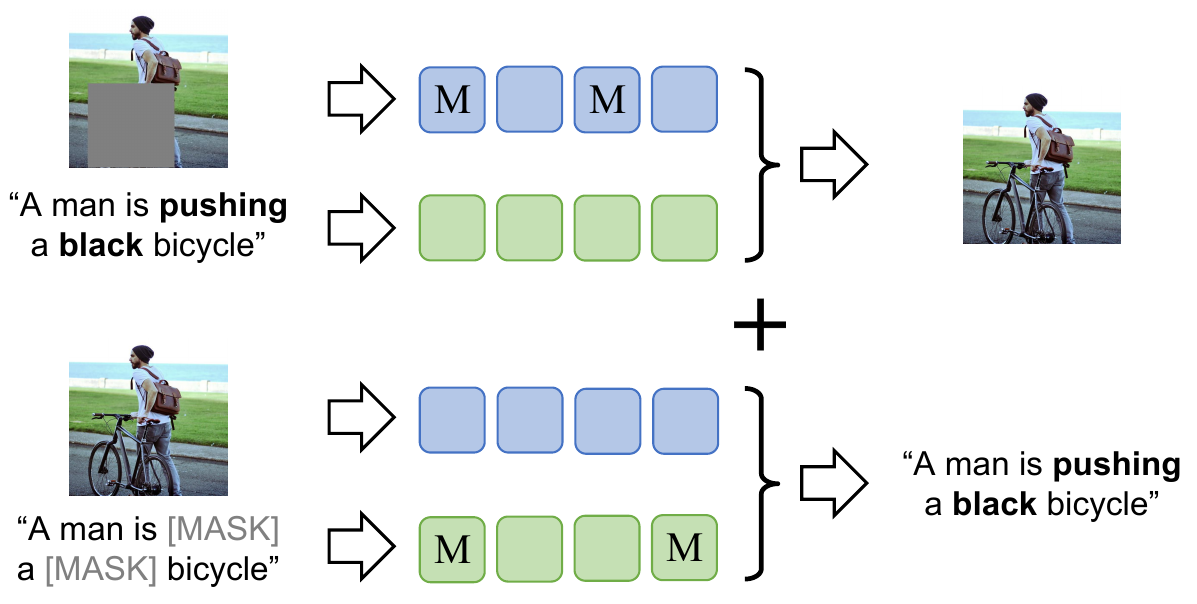}
    \caption{\textbf{The core idea of our method.} We mask compositional concepts in one modality and reconstruct them conditioned on the full information from the other.}
    \label{fig:core_idea}
    \vspace{-6mm}
\end{figure}

However, compositional understanding remains a key limitation. These models often struggle with object relations, attribute-object bindings, and word order dependencies—frequently exhibiting ``bag-of-words'' behavior \cite{yuksekgonul2023when,thrush2022winoground,zhao2022vl,hsieh2024sugarcrepe}. For instance, they tend to fails to distinguish between ``the horse is eating grass'' and ``the grass is eating the horse'' or between ``a black dog with a white cat'' and ``a white dog with a black cat''. Addressing this challenge is crucial for improving VLMs reasoning and facilitating their application in downstream tasks.

To enhance the compositional understanding capabilities of VLMs, most existing approaches focus on the careful construction of hard negative samples with subtle semantic variations, using rule-based templates \cite{yuksekgonul2023when}, LLM-generated captions \cite{doveh2023dense}, or synthetic scenes \cite{cascante2023going}. While effective, these methods are often costly, noisy, and lead the model to focus on superficial patterns specific to those negatives \cite{hsieh2024sugarcrepe, geirhos2020shortcut}. Moreover, recent work \cite{kamath2024hard} shows that reliance on hard negatives may induce oversensitivity, causing models to rank semantically equivalent captions incorrectly. This motivates an intriguing question: \textbf{\textit{Beyond hard negative mining, can we improve compositionality of VLMs by designing a training framework that better exploits the rich aligned compositional information inherently present in existing image-text pairs?}}

In this work, we introduce \textbf{MACCO} (\textbf{MA}sked \textbf{C}ompositional \textbf{C}oncept M\textbf{O}deling), a novel framework that enhances compositionality in VLMs without explicit hard negative construction. Our method masks compositional concepts in one modality and reconstructs it conditionally using full context from the other. As illustrated in Figure \ref{fig:core_idea}, the masked text and full image are used to reconstruct compositional concept words, while the masked image and full text are used to reconstruct image regions corresponding to compositional concepts. To better constrain global features during reconstruction and enrich local tokens with contextual global semantics, we introduce a parameter-free \textit{global-to-local semantic injection} operation.

To facilitate this masked cross-modal reconstruction, we introduce two novel auxiliary objectives. First, the \textbf{Masked-augmented Cross-Modal Alignment Loss (MCA)} integrates global features of masked texts or masked images into the cross-modal contrastive learning process. Second, the \textbf{Masked-augmented Intra-Modal Regularization Loss (MIR)} regularizes the global features of masked instances within each modality to prevent representational collapse. Extensive experiments across five compositional benchmarks and four backbones demonstrate the effectiveness of our approach. In-depth analyses show that MACCO also enhances the model’s ability to capture syntactic structure and semantic nuance. It produces more concept-aware embeddings, exhibits stronger robustness to semantically invariant perturbations, and better preserves fine-grained linguistic information. Moreover, MACCO can be integrated with hard negative mining methods to obtain additional gains. Finally, further experiments show that the improved compositionality also benefits text-to-image generation and multimodal large language models. 

To summarize, our main contributions are:
\vspace{-2mm}
\begin{enumerate}
\item  We introduce a novel framework that improves vision-language compositionality in pre-trained VLMs without requiring explicit hard negative samples, and we show that the improved compositionality also benefits other multimodal tasks.
\item  We propose two auxiliary objectives, MCA and MIR, to promote effective cross-modal reconstruction and alignment learning.
\vspace{-2mm}
\item  We validate the effectiveness of our approach through extensive experiments on five widely used vision-language compositional benchmarks, complemented by in-depth analyses. Our framework is also compatible with existing hard negative mining methods, yielding additional gains when integrated.
\end{enumerate}
\vspace{-5mm}
\begin{figure*}
    \centering
    \includegraphics[width=0.95\linewidth]{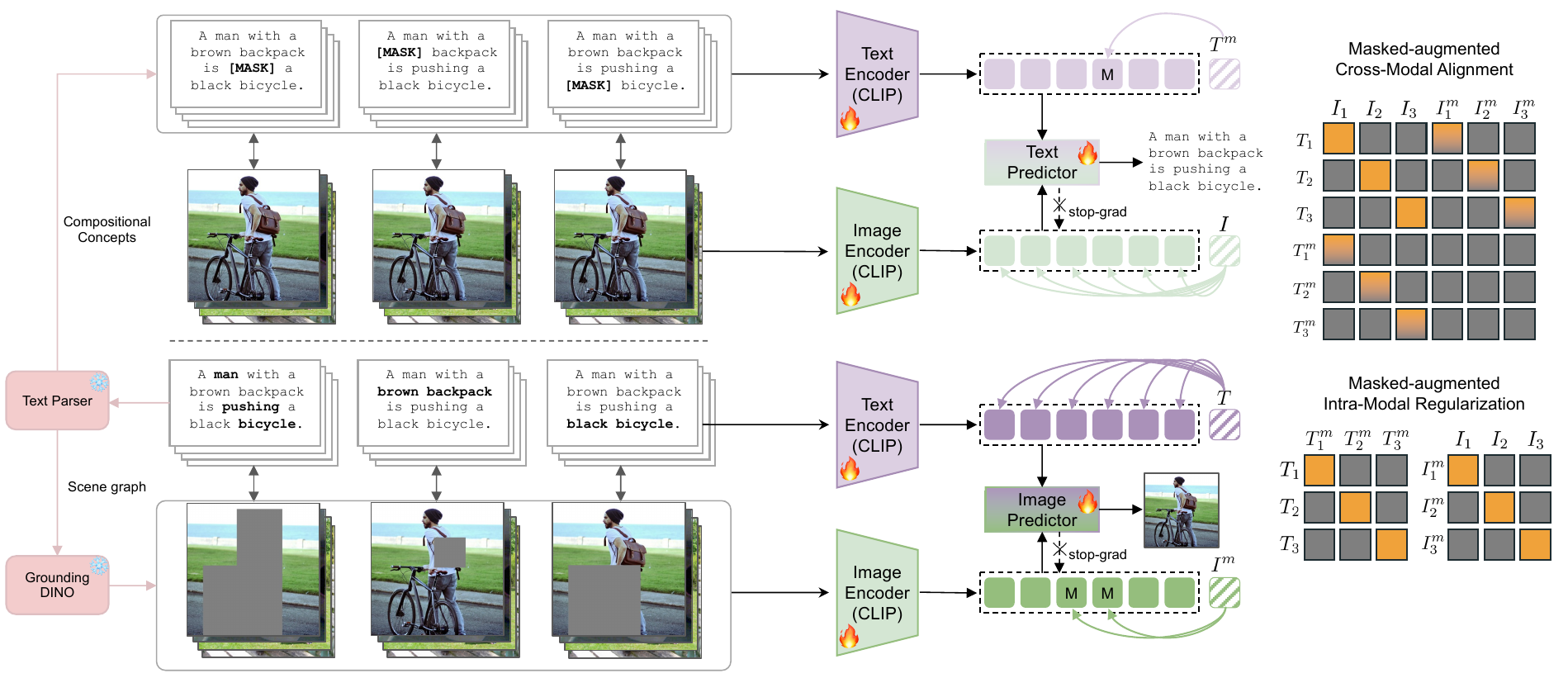}
    \caption{Our framework employs image and text predictors exclusively during training, removing them at inference time. The two image encoders share weights and function as a single encoder, as do the two text encoders.}
    \label{fig:framework}
    \vspace{-4mm}
\end{figure*}

\section{Related Works}
\label{sec:related}
\vspace{-2mm}
\noindent\textbf{Contrastive Vision-Language Models.} Vision-language foundation models have achieved remarkable progress. Representative models such as CLIP \cite{radford2021learning}, pretrained via contrastive learning on large-scale and noisy image-text datasets, exhibit impressive zero-shot transfer capabilities, leading to success across a wide range of tasks. Our motivation to focus on CLIP is twofold. First, contrastive learning has become a dominant and highly effective paradigm for multimodal representation learning. Second, CLIP-like models serve as the foundation of numerous applications, showcasing wide applicability across diverse domains. Enhancing CLIP is therefore of significant value, as improvements can benefits to a broader range of vision-language applications.

\noindent\textbf{Vision-Language Compositionality.}
Despite significant progress, vision-language models such as CLIP still struggle with compositional reasoning—understanding fine-grained relations, attributes, and word order beyond object recognition \cite{yuksekgonul2023when, hsieh2024sugarcrepe, zhao2022vl, thrush2022winoground}. 
To enhance compositionality, most prior work focuses on fine-tuning with hard negative samples, using strategies such as rule-based construction \cite{yuksekgonul2023when}, LLM-based synthesis \cite{doveh2023dense}, and negative image synthesis via diffusion models \cite{li2025enhancing}. Beyond these, SDS-CLIP \cite{basu2024distilling} introduces a novel distillation loss from Stable Diffusion to improve the compositionality of CLIP. CLIP-CAE \cite{li2024interpretable} enhances the model’s attention to compositional concepts by explicitly optimize internal attribution.

\noindent\textbf{Masked Signal Modeling.}
Masked reconstruction is a widely adopted pretraining strategy. In NLP, BERT \cite{devlin2019bert} demonstrates the success of masked language modeling (MLM). Inspired by this, masked image modeling methods (MIM) like BEiT \cite{bao2021beit}, MAE \cite{he2022masked}, and SimMIM \cite{xie2022simmim} train vision transformers to recover masked visual content. In VLMs, MaskVLM \cite{kwon2022masked} jointly reconstructs randomly masked image and text inputs, while \citet{arici2021mlim} explores MIM and MLM for structured catalog data to facilitate downstream vision tasks. These methods highlight the potential of masked modeling for cross-modal representations learning.
\vspace{-2mm}

\section{Method}
\label{sec:method}
\vspace{-2mm}
\subsection{Preliminaries of CLIP}
\vspace{-1mm}
\label{sec:clip_preliminary}

CLIP consists of an image encoder $E_I$ and a text encoder $E_T$, which project images and texts into a shared embedding space. The image encoder produces patch-level features and a global CLS token via full attention, while the text encoder generates token-level representations via causal attention, with the CLS token derived from the EOS token. Given a batch of paired samples $\mathcal{B} = {(I_i, T_i)}_{i=1}^{N}$, CLIP computes the similarity between global image and text embeddings using cosine similarity and is trained via a symmetric InfoNCE loss to align matching pairs and contrast mismatched ones. Detailed formulations are provided in
Appendix \ref{appendix:clip_details}.
\vspace{-5mm}

\subsection{MACCO Framework}
\vspace{-1mm}
\label{sec:framework}
As illustrated in Figure \ref{fig:framework}, our framework enhances compositional understanding by masking compositional concepts in one modality and reconstructing them using the full features of the other modality as context. This design better exploits the aligned compositional signals inherent in paired image-text data. Specifically, masked texts are reconstructed using complete image features, and vice versa for masked images.

Prior studies \cite{kamath2023text,dumpala2024seeing} indicate that contrastive VLMs often struggle with compositional semantics due to limitations in the text encoder, particularly in capturing object relations and attribute bindings. Motivated by this, our framework emphasizes improving the text encoder’s capacity to understand and represent compositional concepts. During reconstruction, we \textit{stop the gradients} of the image features in both predictors, ensuring the loss focuses on optimizing text representations. We analyze this design choice and its impact in Section \ref{sec:ablation} and Appendix \ref{app:ablation_not_freeze_image_encoder}.

\noindent\textbf{Compositional Concept Extraction.} To identify compositional concepts in both text and image modalities, we extract relation and attribute phrases from training examples. For textual inputs, we apply a scene graph parser \cite{wu2019unified} to obtain a mask $M^T$ indicating the token positions of compositional phrases. For visual inputs, we use GroundingDINO \cite{liu2024grounding} to localize image regions corresponding to compositional phrases and map them to CLIP patch indices, producing a mask $M^I$ over visual tokens. The full extraction pipeline and alignment details are provided in Appendix~\ref{app:comp_extraction} and Algorithm~\ref{alg:comp_extraction}.

\noindent \textbf{Feature Extraction Formulation.}
Given a batch $\mathcal{B} = {(I_i, T_i)}_{i=1}^{N}$ of image-text pairs, we first obtain token-level compositional concept masks $\mathcal{M}^T \in \text{bool}^{N \times L}$ and $\mathcal{M}^I \in \text{bool}^{N \times (P+1)}$ for the text and image modalities, respectively, where $L$ is the maximum text length and $P$ is the number of image patches. A value of \texttt{True} indicates that the token is masked. For each modality, we initialize a shared learnable mask token ($m_t$ for text and $m_i$ for image). We then replace the tokens at masked positions with the corresponding mask token and add positional embeddings $PE^T$ and $PE^I$:

\vspace{-4mm}
{
\small
\begin{equation}
    X^T = \text{Embed}(T) + PE^T, X^T_m = \text{Mask}(\text{Embed}(T)) + PE^T,
\end{equation}
}
\vspace{-1mm}
{
\small
\begin{equation}
    X^I = \text{Embed}(I) + PE^I,X^I_m = \text{Mask}(\text{Embed}(I)) + PE^I.
\end{equation}
}

Then, we feed both the masked and unmasked sequences into the respective encoders. The masked and unmasked representations are then encoded as:

\vspace{-2mm}
{
\small
\begin{equation}
    f^T = E_T(X^T),
    f^T_{m} = E_T(X^T_{m}),
\end{equation}
}
\vspace{-2mm}
{
\small
\begin{equation}
    f^I = E_I(X^I),
    f^I_{m} = E_I(X^I_{m}).
\end{equation}
}

Here, $f^T$ and $f^I$ denote the full text features and full image features, while $f^T_m$ and $f^I_m$ correspond to masked variants.

\noindent\textbf{Masked Textual Compositional Concept Modeling.} To reconstruct masked text from cross-modal signals, we extract global features of the masked text ($f_m^{T|cls}$) and the full image ($f^{I|cls}$). Due to the causal nature of the text encoder, masked tokens lack sufficient future context. To mitigate this, we apply a simple \textbf{\textit{global-to-local semantic injection}} operation, in which each masked token is enriched by integrating its representation with the global feature of the masked text, thereby enhancing contextual reasoning within the same modality.

For the image features, since CLIP’s pretraining does not explicitly constrain local patch tokens, their alignment with text is weaker than that of the CLS token \cite{bica2024improving}. Thus, we also inject global semantics into each image patch token to compensate for CLIP’s weak local supervision and strengthen grounding during reconstruction. For simplicity, we formalize the \textit{global-to-local semantic injection} operation as follows:

\vspace{-1.5mm}
{
\setlength{\abovedisplayskip}{0.5pt}
\setlength{\belowdisplayskip}{2.0pt}
\small
\begin{equation}
\bar{f_m^T} = \frac{1}{2}(f_m^T + f_m^{T|cls}),
\bar{f^I} = \frac{1}{2}(f^I + f^{I|cls}).
\end{equation}
}
\vspace{-3.5mm}

The text predictor $D^T$ uses two layers of cross-attention, attending from contextual masked text tokens to full image features, followed by a classification head, which is used to predict in the vocabulary space, following BERT \cite{devlin2019bert}. The final loss for the masked modeling of texual compositional concepts is formulated as:

\vspace{-2mm}
{
\setlength{\abovedisplayskip}{2.5pt}
\setlength{\belowdisplayskip}{2.5pt}
\small
\begin{equation}
    L_{MLM} = \mathbf{E}_{(T,I)\sim D} \mathcal{H}[D^T(\bar{f_m^{T}}, \texttt{stopgrad}(\bar{f^{I}})), T],
\end{equation}
}where $\mathcal{H}$ denotes cross entropy loss. We compute the loss only on masked token.

\noindent \textbf{Masked Visual Compositional Concept Modeling.} For cross-modal image reconstruction, we also apply \textbf{\textit{global-to-local semantic injection}} operation to enrich local tokens with contextual global semantics. Specifically, each local text token is fused with the global text feature $f^{T|cls}$ to obtain $\bar{f^T}$, and the global masked image feature $f_m^{I|cls}$ is similarly injected into local patch tokens to obtain $\bar{f_m^I}$:

\vspace{-2mm}
{
\setlength{\abovedisplayskip}{0.5pt}
\setlength{\belowdisplayskip}{0.5pt}
\small
\begin{equation}
\bar{f_m^I} = \frac{1}{2}(f_m^I + f_m^{I|cls}),\quad \bar{f^T} = \frac{1}{2}(f^T + f^{T|cls}).
\end{equation}
}

The image predictor $D^I$ employs masked image tokens as queries and full text features as keys/values in a three-layer cross-attention module. Following MAE \cite{he2022masked}, we use a decode embedding layer and 2D positional embedding prior to attention. The final prediction head reconstructs pixel values for each patch. The loss is the mean squared error (MSE) between reconstructed and original pixels:

\vspace{-2mm}
{
\setlength{\abovedisplayskip}{3pt}
\setlength{\belowdisplayskip}{3pt}
\small
\begin{equation}
    L_{MIM} = \mathbf{E}_{(T,I)\sim D} \, \|[D^I(\texttt{stopgrad}(\bar{f_m^{I}}), \bar{f^{T}}), I\|^2,
\end{equation}
}
we compute the loss only on the masked patches.

\noindent\textbf{Masked-augmented Cross-Modal Alignment.}
We extend the standard contrastive learning framework in CLIP by incorporating the CLS token features of masked text or image inputs into the contrastive objective. Compared to their complete counterparts, masked inputs lack certain compositional concepts. For example, a masked text may retain only object-level information and thus can serve as a \textit{soft negative sample} in image-to-text contrastive learning. Despite missing some details, the CLS token of masked input is still encouraged to encodes meaningful global semantics, as it facilitates reconstruction through \textit{global-to-local semantic injection}. Thus, to better constrain the semantics, we introduce masked-augmented cross-modal alignment losses by computing contrastive losses between masked text and full images, and vice versa. During alignment, masked inputs are treated as \textit{soft negatives} in the corresponding contrastive objectives. The masked-augmented image-to-text contrastive loss is formulated as follows:

\vspace{-5mm}
{
\small
\begin{equation}
    \begin{split}
     L_{i2t}^{MCA} = \sum\limits_{i=1}^{N} \log \frac{\exp^{S(I_i, T_i)}}{\sum\limits_{j=1}^{N} \exp^{S(I_i, T_j)} + \sum\limits_{j=1}^{N} \exp^{S(I_i, T_j^m)}} \\ + 
     \sum\limits_{i=1}^{N} \log \frac{\exp^{S(I_i^m, T_i)}}{\sum\limits_{j=1}^{N} \exp^{S(I_i^m, T_j)} + \sum\limits_{j=1}^{N} \exp^{S(I_i^m, T_j^m)}}.
     \end{split}
\end{equation}
}
\vspace{-2mm}

Similarly, the masked-augmented text-to-image contrastive loss is formulated as:
{
\small
\begin{equation}
    \begin{split}
    L_{t2i}^{MCA} = \sum\limits_{i=1}^{N} \log \frac{\exp^{S(T_i, I_i)}}{\sum\limits_{j=1}^{N} \exp^{S(T_i, I_j)} + \sum\limits_{j=1}^{N} \exp^{S(T_i, I_j^m)}} \\ +
    \sum_{i=1}^{N} \log \frac{\exp^{S(T_i^m, I_i})}{\sum\limits_{j=1}^{N} \exp^{S(T_i^m, I_j)} + \sum\limits_{j=1}^{N} \exp^{S(T_i^m, I_j^m)}}.
    \end{split}
\end{equation}
}

Finally, the total masked-augmented cross modal contrastive loss is the sum of both:

{
\small
\begin{equation}
   L_{MCA} = - \frac{1}{2}(L_{i2t}^{MCA} + L_{t2i}^{MCA}).
\end{equation}
}

\noindent\textbf{Masked-augmented Intra-Modal Regularization.} To prevent the masked text features or masked image features of different samples from collapsing into the same subspace and to constraint the deviation of the masked features from their corresponding full text or image features, contrastive loss is is well-suited for this regularization purpose. Additionally, contrasting single modality when performing cross-modal alignment is helpful for stable training \cite{zhang2024contrasting}. Therefore, we introduce a new intra-modal regularization loss. Specifically, we apply intra-modal contrastive learning between the masked text features and the full text features, as well as between the masked image features and the full image features. The masked text to original text contrastive loss is formulated as follows:

\vspace{-5mm}
{
\setlength{\abovedisplayskip}{6pt}
\setlength{\belowdisplayskip}{3pt}
\small
\begin{equation}
    \begin{split}
        L_{t2t}^{MIR} = \sum\limits_{i=1}^{N} \left[\log \frac{\exp^{S(T_i^m, T_i)}}{\sum\limits_{j=1}^{N} \exp^{S(T_i^m, T_j)}} + \log \frac{\exp^{S(T_i, T_i^m)}}{\sum\limits_{j=1}^{N} \exp^{S(T_i, T_j^m)}} \right].
    \end{split}
\end{equation}
}

Similarly, the masked image to original image contrastive loss is formulated as:

\vspace{-4mm}
{
\setlength{\abovedisplayskip}{6pt}
\setlength{\belowdisplayskip}{3pt}
\small
\begin{equation}
    \begin{split}
        L_{i2i}^{MIR} = \sum\limits_{i=1}^{N} \left[ \log \frac{\exp^{S(I_i^m, I_i)}}{\sum\limits_{j=1}^{N} \exp^{S(I_i^m, I_j)}} + \log \frac{\exp^{S(I_i, I_i^m)}}{\sum\limits_{j=1}^{N} \exp^{S(I_i, I_j^m)}} \right].
    \end{split}
\end{equation}
}

Finally, the total masked-augmented intra-modal contrastive loss is the sum of both:

\vspace{-2mm}
{
\small
\begin{equation}
   L_{MIR} = -\frac{1}{2}(L_{t2t}^{MIR} + L_{i2i}^{MIR}).
\end{equation}
}

\noindent\textbf{Overall Training Objective.}
Our MACCO incorporates two masked modeling losses, $L_{MLM}$ and $L_{MIM}$, as well as two masked-augmented auxiliary losses, $L_{MCA}$ and $L_{MIR}$. The final loss is formulated as follows:

\vspace{-3.5mm}
{
\setlength{\abovedisplayskip}{6pt}
\setlength{\belowdisplayskip}{3pt}
\small
\begin{equation}
    L_{total} = L_{MCA} + \lambda_{1} L_{MIR} + \lambda_{2} L_{MLM} + \lambda_{3} L_{MIM},
\end{equation}
}where $\lambda_{1}$, $\lambda_{2}$, and $\lambda_{3}$ are the weighting factors for the respective losses.
\vspace{-2mm}
\section{Experiments}
\label{sec:exp}
\begin{table*}[tbp]
    \centering
    \resizebox{\linewidth}{!}{
    \begin{tabular}{l ccc cc cc c c}
            \toprule[1.25pt]
            \multirow{2}{*}{Model} &\multicolumn{3}{c}{ARO} 
            &\multicolumn{2}{c}{Sugar-Crepe} 
            &\multicolumn{2}{c}{VL-Checklist} 
            &\multicolumn{1}{c}{VALSE}
            &\multicolumn{1}{c}{What's-up} 
            \\
            \cmidrule(rrr){2-4} \cmidrule(rr){5-6} \cmidrule(rr){7-8} \cmidrule(r){9-9} \cmidrule(r){10-10}
             & Relation & Attribute & Order & Relation & Attribute & Relation & Attribute & Relation & Relation   \\
            \midrule
            Random Chance &50.0 &50.0 & 20.0 & 50.0  & 50.0 & 50.0 & 50.0 & 50.0 & 41.7 \\
            \midrule
            CLIP$^1$ (ViT-B/32)  & 58.7 & 62.7 & \underline{54.1} & 68.8 & 70.8 & 63.6 & 67.7 & \underline{70.1} & 41.8\\
            CLIP-FT  & 64.3 & \underline{66.2} & 49.1 & 71.1 & 77.7 & 60.9 & 67.4 & 69.3 & 41.4 \\
            IL-CLIP$^2$ & 50.0 & 55.3 & 16.7 & 56.3 & 63.9 & 55.7 & 59.5 & 55.7 & \underline{42.3} \\
            SDS-CLIP$^3$ & 53.0 & 62.0  & 29.0 & - & - & - & - & - & -\\
            CLIP-CAE$^4$ & \underline{69.5} & 65.4 & - & \underline{73.0} & \underline{78.9} & \underline{65.4} & \underline{68.6} & 68.8 & - \\
            \midrule
            \rowcolor{lightgray}
            \textbf{MACCO-CLIP (ours)} & \textbf{73.1} & \textbf{68.5} & \textbf{76.0} & \textbf{77.1} & \textbf{79.1} &  \textbf{70.2} & \textbf{68.7} & \textbf{75.3} & \textbf{43.2} \\
            \bottomrule[1.25pt]
            \multicolumn{10}{l}{\small
            References: $^1$\cite{radford2021learning} $^2$\cite{zheng2024iterated} $^3$\cite{basu2024distilling} $^4$\cite{li2024interpretable}} \\
    \end{tabular}
    }
    \vspace{-3mm}
    \caption{
    \textbf{Results on ARO, SugarCrepe, VL-Checklist, VALSE, and What's-up.} The best results are marked in \textbf{bold}, and the second-best results are \underline{underlined}. Empty entries denote that the model's code has not been released. The result reported for CLIP-CAE are the average performance across its four model instances. The detailed results can be found in Appendix \ref{app:subset_results}.}
    \vspace{-5mm}
    \label{tab:main_results}
\end{table*}
\vspace{-2mm}
\noindent\textbf{Training Setup.} Following \cite{li2024interpretable} and \cite{basu2024distilling}, we use approximately $110$k high-quality image text pairs from MSCOCO \cite{lin2014microsoft} as the training set, and include additional experiments on CC3M in Appendix~\ref{app:exp_beyond_coco}. In the main experiments, we use the widely adopted OpenAI CLIP ViT/B-32 model, and provide supplementary results with ViT/B-16 and ViT/L-14 in Appendix~\ref{app:other_model_scale}. We initialize both the text encoder and the image encoder with pretrained CLIP weights. The image and text predictors are trained from scratch. 
Following previous approaches \cite{basu2024distilling,li2024interpretable}, we fine-tune the model for $5$ epochs with a batch size of $256$ and conduct a $50$ steps warm up. The learning rate for the CLIP model is set to $5$e-$7$, while the learning rate for the two predictors is set to $1$e-$3$. We use AdamW as the optimizer with a weight decay of $0.2$. Experiments are conducted on a single NVIDIA A100 GPU.

\noindent\textbf{Evaluation Setup.} At inference time, both predictors are removed, and the model's architecture remain the same as the pre-trained CLIP model. We perform a comprehensive evaluation using five widely used benchmarks for vision-language compositional understanding: ARO \cite{yuksekgonul2023when}, SugarCrepe \cite{hsieh2024sugarcrepe}, VL-Checklist \cite{zhao2022vl}, VALSE \cite{parcalabescu2021valse} and What's-up \cite{kamath2023s}. Detailed information about these datasets can be found in Appendix \ref{app:comp_benchmark}.

For a fair comparison and comprehensive evaluation, we mainly selected three types of baselines: (i) the pre-trained CLIP model; (ii) the CLIP model fine-tuned on MSCOCO using only the contrastive loss (denoted as CLIP-FT); and (iii) CLIP-CAE \cite{li2024interpretable} and SDS-CLIP \cite{basu2024distilling}, which enhance the compositionality of CLIP-like model through fine-tuning on MSCOCO.
\vspace{-2mm}

\subsection{Main Results}
\vspace{-2mm}
\label{sec:main_results}
As shown in Table~\ref{tab:main_results}, MACCO-CLIP achieves state-of-the-art performance across five widely-used benchmarks, significantly outperforming both the pretrained CLIP model and several fine-tuned variants, including CLIP-FT and CLIP-CAE. These results demonstrate MACCO-CLIP’s strong advantages in relation understanding, attribute binding, and word order sensitivity.

Compared to CLIP, our model yields notable improvements, including $14.4\%$ on ARO-Relation, $5.8\%$ on ARO-Attribute, $21.9\%$ on ARO-Order, and $8.3\%$ average gain on Sugar-Crepe. Against CLIP-FT, we observe gains of $8.8\%$ (ARO-Relation), $6.0\%$ (Sugar-Crepe Relation), and $9.3\%$ (VL-Checklist Relation). Notably, MACCO-CLIP achieves a $26.9\%$ improvement on ARO-Order over CLIP-FT, significantly mitigating the well-documented insensitivity of CLIP to word order.

MACCO-CLIP also consistently outperforms CLIP-CAE across all benchmarks, for example, with gains of $4.1\%$ on Sugar-Crepe Relation and $4.8\%$ on VL-Checklist Relation. While CLIP-CAE underperforms CLIP-FT on ARO-Attribute, our model improves over CLIP-FT by $2.3\%$. These gains may stem from a key design difference: although CLIP-CAE encourages models to focus on compositional concepts, it lacks an explicit mechanism for modeling dependencies between entities and their corresponding relations or attributes. In contrast, MACCO-CLIP incorporates a cross-modal masked modeling objective that explicitly encourage the model to capture such dependencies, resulting in semantically richer and more syntactically coherent representations. We also valid this effect in Section \ref{sec:analysis}.

We note that both our MACCO-CLIP and prior work CLIP-CAE achieve smaller gains on attribute-focused tasks compared to relation-based tasks. This aligns with findings from prior work \cite{huang2023t2i, lewis2024does}, which suggests that attribute binding remains a more challenging aspect of compositional understanding and merits more investigation. Further discussion of this challenge can be found in Appendix~\ref{app:attribute_binding_discussion}.

Finally, we also conduct additional experiments on three models with different scales and training paradigms, namely ViT-B/16, ViT-L/14, and SigLIP, and observe consistent and significant improvements across all of them. Detailed results are presented in Table~\ref{tab:other_model_scale} in Appendix \ref{app:other_model_scale}. Overall, these results highlight the effectiveness of our method.
\vspace{-2mm}

\subsection{Combined with Hard-Negative Samples}
\vspace{-1.5mm}
\label{sec:combined_with_hard_negative}
\begin{table*}[t!]
    \centering
    \setlength{\tabcolsep}{3pt}
    \resizebox{\linewidth}{!}{
    \begin{tabular}{l ccc cc cc c c}
            \toprule[1.25pt]
            \multirow{2}{*}{Model} &\multicolumn{3}{c}{ARO} 
            &\multicolumn{2}{c}{Sugar-Crepe} 
            &\multicolumn{2}{c}{VL-Checklist} 
            &\multicolumn{1}{c}{VALSE}
            &\multicolumn{1}{c}{What's-up} 
            \\
            \cmidrule(rrr){2-4} \cmidrule(rr){5-6} \cmidrule(rr){7-8} \cmidrule(r){9-9} \cmidrule(r){10-10}
             & Relation & Attribute & Order & Relation & Attribute & Relation & Attribute & Relation & Relation   \\
            \midrule
            Random Chance &50.0 &50.0 & 20.0 & 50.0  & 50.0 & 50.0 & 50.0 & 50.0 & 41.7\\
            \midrule  
            CLIP$^1$ (ViT-B/32)  & 58.7 & 62.7 & 54.1 & 68.8 & 70.8 & 63.6 & 67.7 & 70.1 & 41.8 \\
            \midrule
            NegCLIP$^2$ & 80.4 & 71.7 & 91.7 & 73.2 & 80.0 & 71.8 & 70.1 & 79.5 & 42.1\\
            \rowcolor{lightgray}
            \textbf{+\ MACCO}  & \underline{80.2} & \underline{71.6} & \textbf{92.1} & \textbf{74.9} & \underline{79.7} & \textbf{74.9} & \textbf{70.6} & \underline{78.7} & \textbf{43.1} \\
            \midrule
            CE-CLIP$^3$ & 82.2 & 72.9 & 95.1 & 72.6 & 80.1 & 75.6 & 69.4 & 78.5 & 44.4 \\
            \rowcolor{lightgray}
            \textbf{+\ MACCO}  & \textbf{82.6} & \textbf{73.0} & \textbf{96.4} & \textbf{72.7} & \underline{80.0} & \textbf{77.5} & \textbf{70.5} & \textbf{79.2} & \textbf{44.7}\\
            \bottomrule[1.25pt]       
            \multicolumn{10}{l}{\small
            References: $^1$\cite{radford2021learning} $^2$\cite{yuksekgonul2023when} $^3$\cite{zhang2024contrasting}}
    \end{tabular}
    }
    \vspace{-3mm}
    \caption{
    \textbf{Results on ARO, Sugar-Crepe, VL-Checklist, VALSE and What's-up when combined with hard negative samples}. Highlighted in \textbf{bold} denote an improvement over NegCLIP or CE-CLIP, while the \underline{underlined} ones indicate a performance degradation compared to NegCLIP or CE-CLIP.}
    \label{tab:combined_with_hn}
    \vspace{-3mm}
\end{table*}
Since our framework is orthogonal to hard negative mining approaches, we further investigate whether it can be effectively integrated with hard negative samples. We consider two representative methods based on hard-negative samples: NegCLIP \cite{yuksekgonul2023when}, which incorporates hard negatives within a standard contrastive learning framework, and CE-CLIP \cite{zhang2024contrasting}, which introduces two additional contrastive loss terms to better leverage the hard negatives.

For fair comparison, we use the same hard negative samples provided by NegCLIP across all experiments. As shown in Table \ref{tab:combined_with_hn}, without bells and whistles, both models consistently benefit from the integration with MACCO, with the most notable improvements observed on VL-Checklist. For instance, MACCO enhances the performance of NegCLIP by $3.1\%$ and CE-CLIP by $1.9\%$ on the VL-Checklist Relation. These results further highlight the effectiveness of our method and its plug-and-play compatibility with existing hard negative mining methods. In Appendix \ref{app:discussion_hard_negative_method}, we provide a more detailed discussion on our approach and hard-negative mining methods.
\vspace{-2.5mm}

\subsection{Downstream Tasks}
\vspace{-1.5mm}
\label{sec:downstream_task}
In Table~\ref{tab:downstream_task}, we present the zero-shot classification accuracy and linear probing results on $11$ widely used classification benchmarks. Details of the linear probing protocol settings are provided in the Appendix \ref{app:downstream}.
The results show that MACCO-CLIP incurs only a slight reduction in zero-shot and linear probe performance compared to the original CLIP model (a decrease of just $1.5\%$ and $0.4\%$ respectively). These results indicate that our method significantly improves compositional understanding while largely preserve the representation capacity of the original CLIP model. Nevertheless, simultaneously enhancing general representation while improving compositional understanding remains an open question and warrants further investigation.
\begin{table}[tb]
    \centering
    \resizebox{\linewidth}{!}{
    \begin{tabular}{l c c c c c}
            \toprule[1.25pt]
            Model & \makecell[c]{Zero-Shot\\ Avg.} & \makecell[c]{Linear Probe\\ Avg.} & \makecell[c]{Comp.\\ Avg.}
            \\
            \midrule 
            CLIP &  \textbf{59.5} & \textbf{80.1} & 61.2 \\
            CLIP-FT  &  57.9 & {80.0} & 61.8\\
            \rowcolor{lightgray}
            \textbf{MACCO-CLIP (ours)}  &  {58.0} & 79.7 & \textbf{67.7}\\
            \bottomrule[1.25pt]        
    \end{tabular}
    }
    \vspace{-2mm}
    \caption{\textbf{Zero-shot classification performance and linear probe results on 11 datasets.} The results in last column represent the average performance across five compositional understanding benchmarks.}
    \label{tab:downstream_task}
\end{table}
\begin{table}[tbp]
    \centering
    \resizebox{\linewidth}{!}{
    \begin{tabular}{l cc cc}
            \toprule[1.25pt]
            \multirow{2}{*}{Model} &\multicolumn{2}{c}{SICK-R}
            &\multicolumn{2}{c}{STS-Benchmark}
            \\
            \cmidrule(rr){2-3} \cmidrule(rr){4-5}
             & Spearman & Pearson & Spearman & Pearson \\
            \midrule 
            CLIP & 67.9 & 68.6 & 61.5 & 59.1 \\
            \midrule 
            CLIP-FT & 68.0 & 73.4 & {66.3} & 64.0\\
            CLIP-CAE & 69.3 & 71.6 & \textbf{66.5} & \textbf{65.2} \\
            \rowcolor{lightgray}
            \textbf{MACCO-CLIP (ours)} & \textbf{70.5} & \textbf{76.4} & 65.3 & {64.9}\\
            \bottomrule[1.25pt]        
    \end{tabular}
    }
    \vspace{-2mm}
    \caption{\textbf{Semantic textual similarity results on SICK-R and STS-Benchmark.}}
    \label{tab:STS}
\end{table}

\vspace{-13mm}
\subsection{Analysis}
\vspace{-1mm}
\label{sec:analysis}
\noindent\textbf{Semantic Textual Similarity.}
Following prior work CLIP-CAE \cite{li2024interpretable}, we evaluate the text encoders of different models on two widely used STS benchmarks: STS-Benchmark \cite{cer2017semeval} and SICK-R \cite{marelli2014semeval}. Details about the task can be found in Appendix \ref{app:STS_benchmark}.

As shown in Table \ref{tab:STS}, our method achieves a notable improvement on SICK-R over CLIP-FT, and outperforms CLIP-CAE with a $4.8\%$ gain in Pearson correlation. While slightly underperforming on STS-Benchmark, we attribute this to its domain heterogeneity and reliance on shallow lexical cues, where CLIP-CAE’s keyword-focused optimization provides a slight advantage. In contrast, SICK-R demands deeper compositional reasoning and sensitivity to lexical-syntactic structure. These findings highlight that MACCO enhances the text encoder’s ability to capture nuanced semantic and compositional relations, beyond surface similarity.

\noindent\textbf{Linguistic Information Probing.}
The experiments on the STS task demonstrate that our model is more effective at capturing compositional semantic information within sentences. To further examine the linguistic properties encoded in the text embedding produced by different models, we perform a probing analysis using the SentEval toolkit \cite{conneau2018senteval} on the text encoders of different models. We use four representative tasks that evaluate the extent to which sentence embeddings encode latent structural and semantic information.  As shown in Table \ref{tab:senteval}, CLIP-FT shows performance degradation on three tasks relative to the original CLIP, whereas MACCO-CLIP consistently yields substantial accuracy gains across all tasks. These results further confirm that our method not only improves the model’s ability to encode compositional concepts but also enhances the text encoder’s capacity to capture syntactic structure and linguistic information.

\begin{table}[tb]
    \centering
    \resizebox{\linewidth}{!}{
    \begin{tabular}{l c c c c c}
            \toprule[1.25pt]
            Model      & Depth & TopConstituents & BigramShift & Tense & Avg.    \\
            \midrule
            CLIP & 25.0 & 50.5   & 64.8       & 82.4  & 55.7   \\
            \midrule
            CLIP-FT    & 25.5 & 49.1  & 64.5       & 82.3  & 55.4  \\
            \rowcolor{lightgray}
            \textbf{MACCO-CLIP (ours)} & \textbf{26.2} & \textbf{50.7}   & \textbf{65.7}  & \textbf{84.0} & \textbf{56.6} \\
            \bottomrule[1.25pt]        
    \end{tabular}
    }
    \vspace{-2mm}
    \caption{\textbf{Probing results of linguistic information in text embedding.}}
    \label{tab:senteval}
    \vspace{-6mm}
\end{table}

\begin{figure}[tb]
    \centering
    \includegraphics[width=1.0\linewidth]{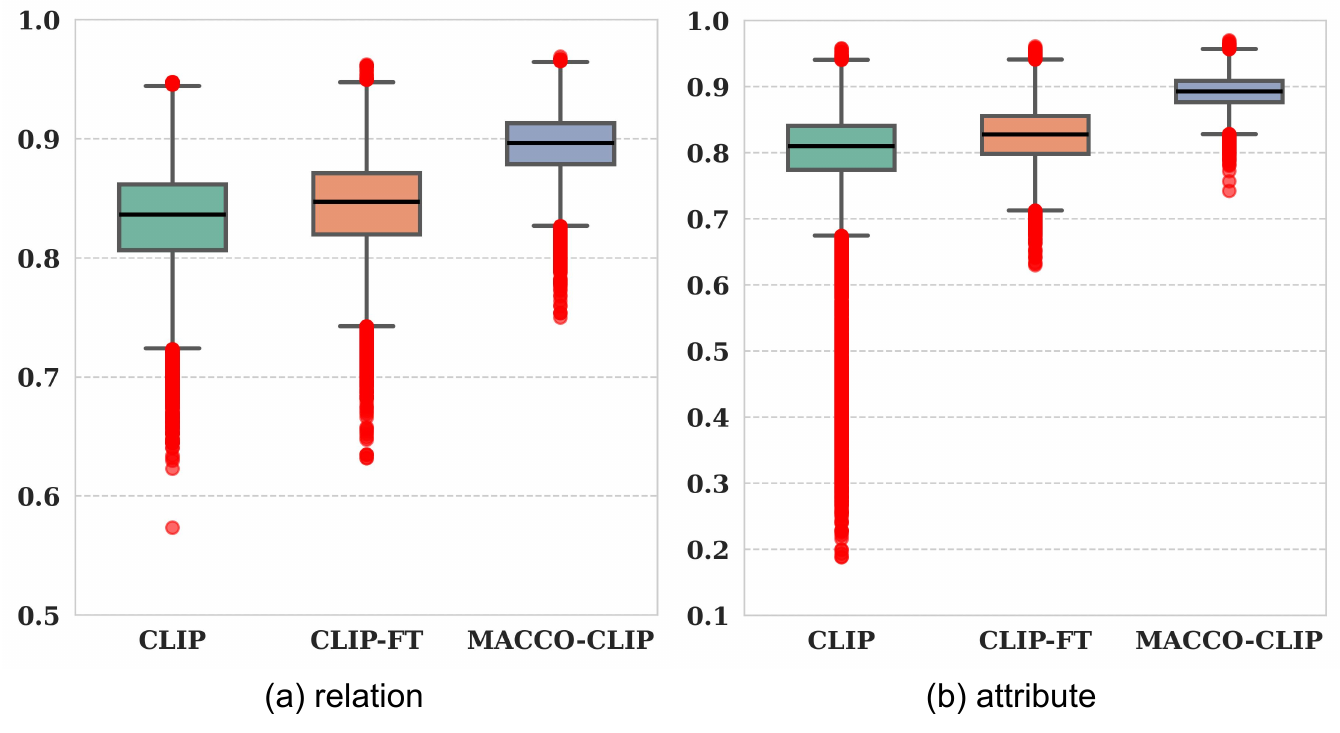}
    \caption{The similarity distribution between the embeddings of full captions and those of relation or attribute phrases extracted from the same text.}
    \label{fig:text_embedding_ingredients}
    \vspace{-2mm}
\end{figure}

\noindent\textbf{Text Embedding Ingredients.}
Inspired by \cite{li2024interpretable}, we follow the procedure outlined in their work to compute the similarity between the embedding of the full caption and that of the corresponding relation or attribute phrase in the ARO benchmarks. Figure \ref{fig:text_embedding_ingredients} presents the similarity distributions for CLIP, CLIP-FT, and our MACCO-CLIP. As shown, the text encoder of MACCO-CLIP produces embeddings that exhibit significantly higher similarity to their corresponding compositional concept embeddings, compared to those generated by CLIP and CLIP-FT. This result further validates that our model more effectively captures compositional concepts in text, with its embeddings encapsulating richer semantic information. 

\noindent\textbf{Compositionality Robustness Evaluation.}
To assess robustness to semantically invariant perturbations, we evaluate MACCO-CLIP on the Hard Positive Compositional Benchmark \cite{kamath2024hard}. The results are shown in Table~\ref{tab:hard_positive_truth}. \textit{Orig. Test Acc.} measures the ability to distinguish positives from meaning-altering hard negatives, while \textit{Aug. Test Acc.} additionally requires recognizing semantically equivalent hard positives. For example, given an image $I$ and captions ``brown grass'' (positive $T$), ``blue grass'' (hard negative $T_n$), and ``chestnut grass'' (hard positive $T_p$), the model must satisfy: $S(I,T)>S(I,T_n)$ and $S(I,T_p) > S(I,T_n)$. 

The results show that MACCO-CLIP consistently outperforms CLIP-FT on both metrics, with notable gain on the \texttt{SWAP} subset. This subset presents a more challenging scenario that tests compositional understanding by constructing hard positives through the reordering of object-attribute phrases. On this subset, MACCO-CLIP achieves a $4.6\%$ improvement over CLIP-FT, highlighting its superior capability in capturing object-attribute binding relationships. These results demonstrate that our method not only improves the model’s sensitivity to meaning-altering perturbations but also enhances its robustness to semantically equivalent variations. This highlights our framework’s strong robustness in visio-linguistics compositionality.
\begin{table}[tbp]
    \centering
    \resizebox{\linewidth}{!}{
    \begin{tabular}{l cc cc}
            \toprule[1.25pt]
            \multirow{2}{*}{Model} &\multicolumn{2}{c}{REPLACE}
            &\multicolumn{2}{c}{SWAP}
            \\
            \cmidrule(rr){2-3} \cmidrule(rr){4-5}
             & \makecell[c]{Orig. \\ Test Acc.} & \makecell[c]{Aug. \\ Test Acc.} & \makecell[c]{Orig. \\ Test Acc.} & \makecell[c]{Aug. \\ Test Acc.} \\
            \midrule 
            CLIP & 63.2 & 47.2 & 61.0 & 49.7\\
            \midrule 
            CLIP-FT & 62.3 &  48.3 &  63.9 & 48.4\\
            \rowcolor{lightgray}
            \textbf{MACCO-CLIP (ours)} & \textbf{68.3} &  \textbf{48.5} &  \textbf{66.1} & \textbf{53.8}\\
            \bottomrule[1.25pt]        
    \end{tabular}
    }
    \vspace{-1.5mm}
    \caption{\textbf{Results on Hard Positive Benchmark.}}
    \label{tab:hard_positive_truth}
    \vspace{-3mm}
\end{table}

\noindent\textbf{Robustness Analysis and Discussion About Detection Model.} In Appendix \ref{app:robust_to_detection_model}, we demonstrate that highly proficient off-the-shelf visual grounding model is not a strict requirement for our method and MACCO is resilient to noisy detection results. We further discuss the scenario generalizability and potential systemic biases of external pre-trained tools in Appendix~\ref{app:discussion_tool_dependency} and Appendix~\ref{app:discussion_tool_bias}. Due to space limitations, more detailed analysis and discussion can be found there. 
\vspace{-2mm}

\subsection{Ablation}
\label{sec:ablation}
We conduct ablation studies to understand the effectiveness of each component in our framework (see Table~\ref{tab:ablation} and Table~\ref{tab:strategy_ablation} in Appendix \ref{app:ablation}). We also provide a clearer ablation of mask strategies and auxiliary objectives in Table~\ref{tab:simplified_ablation}.

\noindent\textbf{Cross-Modal Masked Modeling Losses.} Experimental results show that incorporating reconstruction losses for both text and image improves model performance, regardless of whether auxiliary objectives are included. The best results are achieved when both losses are combined, highlighting the effectiveness of the masked modeling framework.

\noindent\textbf{Auxiliary Losses.} Even without masked modeling, introducing each auxiliary loss individually yields consistent performance gains, achieving the highest improvement when used together. When used with masked modeling, adding $L_{\text{MAC}}$ brings a significant boost, and further incorporating $L_{\text{MIR}}$ leads to the best performance. These results suggest that each auxiliary loss is beneficial on its own, and that compositional masked modeling synergizes with the auxiliary losses to enhance feature representation learning.

\noindent\textbf{Global-to-local Semantic Injection.} The ablation results in Table~\ref{tab:strategy_ablation} and Table~\ref{tab:siglip_g2l_ablation} shows that our method performs better with this strategy, confirming its effectiveness. We attribute this to its ability to enrich local tokens 
with global semantic context and to provide an additional constraint on the global representation.

\noindent\textbf{Stop-Gradient and Masking Strategy.} Blocking gradient flow from image features within the predictors yield better performance, due to a sharper focus on optimizing the text encoder. And our masking strategy targeting compositional concepts clearly outperforms random masking. We provide a more detailed discussion on masking strategies in Appendix \ref{app:masked_modeling_related_works}.


\begin{table*}[htbp]
\centering
\resizebox{1.0\linewidth}{!}{
\begin{tabular}{l  ccc  ccc}
    \toprule[1.25pt]
    \multirow{2}{*}{Model} & \multicolumn{3}{c}{BLIP-VQA $\uparrow$} & \multicolumn{3}{c}{Human-preference $\uparrow$} \\ 
    \cmidrule(lr){2-4} \cmidrule(lr){5-7} 
     & Color & Texture & Shape & Color & Texture & Shape\\
    \midrule
    \makecell[l]{SD 1.5 (w/ vanilla CLIP text encoder)} & 0.3651 & 0.4135 & 0.3721 & -0.4381 & -0.4349 & -0.3323 \\
    \cellcolor{gray!20}{\makecell[l]{\textbf{SD 1.5 (w/ MACCO-CLIP text encoder)}}} & \cellcolor{gray!20}\textbf{0.3815} & \cellcolor{gray!20}\textbf{0.4236} & \cellcolor{gray!20}\textbf{0.3835} & \cellcolor{gray!20}\textbf{-0.3295} & \cellcolor{gray!20}\textbf{-0.3840} & \cellcolor{gray!20}\textbf{-0.2793}\\
    \bottomrule[1.25pt]
    \end{tabular}
    }
    \vspace{-1.5mm}
    \caption{Experimental results on compositional text-to-image generation tasks.}
    \label{tab:application_comp_t2i}
    \vspace{-2mm}
\end{table*}
\begin{table*}[htbp]
\centering
\resizebox{1.0\linewidth}{!}{
\begin{tabular}{l ccccc c}
    \toprule[1.25pt]
    \multirow{2}{*}{Model} & \multicolumn{5}{c}{AMBER} & \multicolumn{1}{c}{MME} \\ 
    \cmidrule(lr){2-6} \cmidrule(lr){7-7} 
    & Attribute & State & Number & Action & Relation & Perception \\
    \midrule
    \makecell[l]{LLaVA-1.5-7B (w/ vanilla CLIP vision encoder)} & 75.8 & 73.9 & 78.2 & 81.1 & 68.4 & 1447.1 \\
    \cellcolor{gray!20}{\makecell[l]{\textbf{LLaVA-1.5-7B (w/ MACCO-CLIP vision encoder)}}} & \cellcolor{gray!20}\textbf{76.5} & \cellcolor{gray!20}\textbf{74.8} & \cellcolor{gray!20}\textbf{78.2} & \cellcolor{gray!20}\textbf{81.7} & \cellcolor{gray!20}\textbf{69.3} & \cellcolor{gray!20}\textbf{1452.3}\\
    \bottomrule[1.25pt]
    \end{tabular}
    }
    \vspace{-1.5mm}
    \caption{Experimental results when applying MACCO's vision encoder to multimodal large language models.}
    \label{tab:application_mllm}
    \vspace{-3mm}
\end{table*}

\subsection{Application}
\vspace{-1mm}
\label{sec:application}
\noindent\textbf{Compositional Text-to-Image Generation.} Text-to-image (T2I) diffusion models such as Stable Diffusion \cite{rombach2022high} typically use pre-trained VLMs (\eg, CLIP) as their text encoders. We therefore investigate whether MACCO, by improving the compositionality of VLMs, also enhances compositional generation in T2I models. As an extended application, we apply the text encoder trained under our MACCO framework to compositional text-to-image generation. Specifically, we replace the original text encoder (ViT-L/14) of Stable Diffusion v1.5 (SD~1.5) with the MACCO-CLIP (ViT-L/14) text encoder. We evaluate attribute binding on T2I-CompBench \cite{huang2023t2i}, which contains three subsets (color, texture, and shape), each with $300$ text prompts. For each prompt, we generate $10$ images with different random seeds and evaluate them using BLIP-VQA scores \cite{huang2023t2i} and ImageReward \cite{xu2023imagereward} preference scores. As shown in Table~\ref{tab:application_comp_t2i}, using the text encoder from MACCO-CLIP improves the attribute binding performance of the T2I model without additional fine-tuning. These results indicate that the stronger text representation backbone learned by MACCO also benefits text-to-image diffusion models and supports more accurate generation under compositional semantics.

\noindent\textbf{Multimodal Large Language Models.} Given that mainstream multimodal large language models (MLLMs) commonly adopt VLMs such as CLIP as visual backbone, we conduct transfer experiments based on the LLaVA-1.5-7B \cite{liu2024improved} to further assess the potential of our method for improving MLLMs. We follow the two-stage training recipe of LLaVA \cite{liu2024improved}, and use LoRA for the instruction-tuning stage due to limited computational resources. We replace the visual encoder of LLaVA-1.5-7B with the vanilla CLIP ViT-L/14 and our MACCO-CLIP ViT-L/14 respectively, and perform the same two-stage training with identical data and hyperparameters. We then evaluate compositional perception performance on AMBER \cite{wang2023amber}, a benchmark for multimodal hallucination, and on MME \cite{fu2025mme}, a general multimodal benchmark. As shown in Table~\ref{tab:application_mllm}, using the visual encoder of MACCO-CLIP improves performance over the baseline across multiple AMBER dimensions, including attributes, states, actions, and relations, and also improves MME perception scores, indicating that the compositional gains from MACCO can also transfer to MLLMs. These results suggest that MACCO strengthens compositional semantic modeling through cross-modal compositional concept masked modeling, thereby enhancing the visual encoder's ability to capture fine-grained visual cues. In contrast to the compositional perception deficiencies of the original CLIP visual encoder \cite{yuksekgonul2023when}, MACCO produces visual representations with richer structural information. 

\vspace{-1mm}

\section{Conclusion}
\label{sec:conclusion}
\vspace{-2mm}
Our work introduced MACCO, a framework that improves compositional understanding in VLMs like CLIP. By masking compositional concepts in one modality and reconstructing them from the other, MACCO better exploits the aligned compositional signals in paired image-text data. We further proposed two auxiliary objectives, MCA and MIR, to enhance cross-modal alignment and intra-modal regularization. Extensive experiments and in-depth analyses show that MACCO effectively improves compositional reasoning, enhances the model’s encoding of syntactic structure and semantic nuance, and benefits other multimodal tasks.

\section{Limitations}
\label{sec:limitation}
While MACCO introduces a novel and effective framework for enhancing the compositional understanding of vision-language models without relying on explicit hard negative construction, several limitations remain, each pointing to promising avenues for future research. \textbf{First}, although MACCO leverages naturally aligned image-text pairs for masked cross-modal reconstruction, it requires lightweight pre-processing to extract compositional concepts (\eg, phrases or regions) from both modalities. This step, while minimal and compatible with standard tools, introduces a dependency that may limit flexibility in fully end-to-end pipelines. \textbf{Second}, MACCO adds two predictors and prediction heads during training, increasing the number of training-time parameters. These components are discarded at inference, but the approach still incurs greater training overhead than methods such as CLIP-CAE \cite{li2024interpretable}. Improving the efficiency and transparency of MACCO’s learned representations remains an important goal. \textbf{Third}, the current design targets contrastive vision-language models like CLIP. Its applicability to generative architectures such as BLIP \cite{li2022blip} has yet to be explored. Extending MACCO to generative objectives, especially those based on language modeling or captioning, is a natural and valuable direction for future work. \textbf{Lastly}, while MACCO enhances compositional robustness and alignment, it does not yet offer the interpretability of concept bottleneck models or attribution enhancement frameworks. Incorporating mechanisms such as concept probing or attributing tracing could yield deeper insights into model behavior. Despite these limitations, MACCO contributes a promising training paradigm for compositional reasoning in VLMs. In addition, MACCO leaves room for further improvement. For example, incorporating multi-granularity alignment as in X-VLM \cite{zeng2022multi} or adopting a stronger text encoder may further enhance MACCO. We provide detailed discussions in Appendix~\ref{app:discussion_with_xvlm} and Appendix~\ref{app:discussion_llm_based_text_encoder}.

\section{Acknowledgements}
\label{sec:acknowledgement}
This work was supported by the Natural Science Foundation of China under Grant 62571507. We sincerely thank the meta-reviewer and the anonymous reviewers for their constructive and valuable feedback.

\bibliography{custom}

\appendix
\label{sec:appendix}
\section{Preliminaries of CLIP}
\label{appendix:clip_details}
CLIP typically consists of two independent encoders: an image encoder $E_I$ and a text encoder $E_T$. Consider a mini-batch $\mathcal{B} = \{(I_i, T_i)\}_{i=1}^{N}$ of size $N$, consisting of image and text pairs $(I_i, T_i)$. The image encoder first divides each image $I_i$ into several image patches, which are embedded into a token sequence, and then positional encoding is added before feeding them into a transformer model. The output is a series of image tokens $V = \{v^{cls}, v^1, \dots, v^P \} \in \mathbb{R}^{(P + 1) \times d}$, where $v^{cls}$ denotes the CLS token that encapsulates global information, $v^i$ represents the patch embedding containing the local information of the image, and $d$ denotes the feature dimension, while $P$ denotes the number of patches. The image encoder employs full attention, meaning all patches and CLS token can attend to each other.

Similarly, the text encoder $E_T$ tokenizes each text $T_i$, then pads it with padding token, adds positional embedding, and feeds it into the transformer model. The output is a series of text tokens $T = \{t^0, t^1, \dots, t^{cls}, \dots, t^{L}\} \in \mathbb{R}^{L \times d}$. $t^{cls}$ is the text-side CLS token, initialized with the EOS token, which encapsulates the global information of the text. The text encoder uses causal attention, meaning that each token can only attend to itself and the previous tokens.

The image-text similarity is measured using the similarity of their global representations:

{
\small
\begin{equation}
    S(I_i, T_j) = \frac{v^{\text{cls}}_i \cdot t^{\text{cls}}_j}{\|v^{\text{cls}}_i\| \cdot \|t^{\text{cls}}_j\|}/\tau,
\end{equation}
}
where $\tau$ is the temperature parameter.

CLIP maximizes the similarity between each matching image-text pair using an InfoNCE loss while minimizing the similarities with other non-matching image-text pairs. The Image-Text Contrastive (ITC) loss is formulated as follows:

{
\setlength{\abovedisplayskip}{2pt}
\setlength{\belowdisplayskip}{2pt}
\small
\begin{equation}
    \mathcal{L}_{ITC} = - \sum\limits_{i=1}^{N} \left[\log \frac{\exp^{S(I_i, T_i)}}{\sum\limits_{j=1}^{N} \exp^{S(I_i, T_j)}} + \log \frac{\exp^{S(T_i, I_i)}}{\sum\limits_{j=1}^{N} \exp^{S(T_i, I_j)}}\right].
\end{equation}
}

\section{Details of Compositional Concept Extraction}
\label{app:comp_extraction}
\noindent\textbf{Textual Compositional Concepts Extraction.} For each text in the training set, we utilize a widely adopted text scene graph parser \cite{wu2019unified} to extract compositional concepts. This parser converts each text into a scene graph by identifying object-relation phrases and object-attribute phrases. It also provides the exact words in the text corresponding to each relation or attribute. For example, the sentence ``A man with a brown backpack is pushing a black bicycle'' will be parsed as: { ``man pushing bicycle'', ``brown backpack'', ``black bicycle''} with compositional concept words being [``pushing'', ``brown'', ``black'']. In this way, we identify the compositional concepts (\ie, relations and attributes) contained in each text. Finally, for each text, we can generate a binary text token mask $M^T$ to indicate the positions of the compositional concepts within the text.

\noindent\textbf{Visual Compositional Concepts Extraction.} For each image in the training set, we leverage the scene graph annotations derived from its caption (\ie, the object-relation and object-attribute phrases), along with a open-set object detector, GroundingDINO \cite{liu2024grounding}. These open-world detection model does not require predefined categories and can detect regions in the image corresponding to the input textual description. Other open-world detection models are also applicable, and empirical results indicate that a reasonably capable detector can yield competitive performance, demonstrating robustness to the choice of detection model. In our main experiments, we use the base version of GroundingDINO and follow the official GitHub repository for inference setup. 

Specifically, we input each relation or attribute phrase from the image caption into the detection model and obtain bounding box coordinates for the matching region. These coordinates are based on the unnormalized coordinates of the original image. We then apply a simple coordinate mapping algorithm to map these coordinates into the image coordinate space used by the CLIP model (since the images input into CLIP often undergo random cropping and resizing as part of data augmentation, we track the parameters of the image data augmentation process to facilitate the coordinate mapping). Next, we further map the region corresponding to these coordinates to specific image patches. Finally, we obtain the positions of the patches in the image corresponding to each relation or attribute phrase. For each relation or attribute phrase, we can generate a binary image token mask $M^I$ to indicate the positions of the compositional concepts within the image corresponding to the phrase. The detailed extraction process is described in Algorithm \ref{alg:comp_extraction}.

\begin{algorithm}[!h]
\caption{Visual Compositional Concepts Extraction}
\label{alg:comp_extraction}
\begin{algorithmic}[1]
\Require Training image $I$, corresponding scene graph phrases $P$, detector $\mathcal{G}$
\Ensure Image token masks of compositional concept \{$M^I(p)$\} for $p \in P$

\State Initialize image preprocessor $\mathcal{C}$ (can track RandomResizedCrop parameter)
\State $\theta \gets (i,j,h,w) \leftarrow \mathcal{C}(I_k)$ \Comment{Record parameter}
\For{each phrase $p \in P_k$}
    \State $B_p \leftarrow \mathcal{G}(I_k, p)$ \Comment{Raw detection}
    \For{each object coordinate $B_p^i$ in $B_p$}
        \State $\hat{B}_p^i \leftarrow \text{CoordinateMapper}(B_p^i, \theta)$ \Comment{Map coordinates into space after apply image preprocessor}
        \State $M^i(p) \leftarrow \text{GenerateMask}(\hat{B}_p^i)$
    \EndFor
    \State Merge image token masks belonging to the sample phrase $p$ to $M^I(p)$
\EndFor
\State
\Function{CoordinateMapper}{$B_p, \theta$}
    \State Parse $\theta = (i,j,h,w)$
    \State Compute scaling factors $(s_w, s_h) \gets (224/w, 224/h)$
    \State Transform coordinates:
    $$
    \hat{B}_p = \begin{bmatrix}
        \max(B_p^{x_1}-j, 0) \cdot s_w \\
        \max(B_p^{y_1}-i, 0) \cdot s_h \\
        \min(B_p^{x_2}-j, w) \cdot s_w \\
        \min(B_p^{y_2}-i, h) \cdot s_h
    \end{bmatrix}
    $$
    \State \Return $\hat{B}_p$
\EndFunction
\State
\Function{GenerateMask}{$\hat{B}_p$}
    \State \Return Image patch indices covered by $\hat{B}_p$
\EndFunction
\end{algorithmic}
\end{algorithm}

\section{Details on Evaluation Benchmark}
\label{app:datasets}
\subsection{Compositionality Benchmark}
\label{app:comp_benchmark}
To comprehensively assess the effectiveness of our method in improving compositional understanding, we conduct evaluations on five widely used compositional benchmarks, as well as a recently proposed benchmark that emphasizes robustness under semantically invariant perturbations. Below, we summarize the details of each dataset.

\textbf{ARO} \cite{yuksekgonul2023when} systematically evaluates vision-language models on three core aspects of compositionality: relations, attributes, and word order. It comprises four major subsets: ARO-Relation, ARO-Attribute, and ARO-Order (which includes both COCO-Order and Flickr30k-Order). ARO-Relation spans $48$ relation types and $23,937$ test samples, requiring models to accurately distinguish relational structures such as ``a dog behind a tree'' versus ``a tree behind a dog''. ARO-Attribute includes $117$ attribute-object combinations across $28,748$ samples, challenging models to resolve attribute compositionality (\eg, distinguishing between ``a crouching cat and an open door'' and ``an open cat and a crouching door''). ARO-Order assesses sensitivity to word order by presenting four permuted versions of a caption, with the model tasked to identify the correct one. Performance is averaged over the COCO-Order and Flickr30k-Order subsets.

\textbf{Sugar-Crepe} \cite{hsieh2024sugarcrepe} is a recently introduced benchmark focused on evaluating models with adversarially generated hard negatives. Leveraging large language models, it produces fluent and semantically plausible negative captions through targeted insertions, replacements, or rephrasings. Following \cite{li2024interpretable}, we report accuracy on the relation and attribute subsets of SugarCrepe separately.

\textbf{VL-Checklist} \cite{zhao2022vl} is a large-scale compositionality evaluation dataset composed of over $410,000$ samples sourced from VG, SWIG, VAW, and HAKE. It covers a wide array of subcategories including color, material, size, action, and spatial relations. Consistent with prior work, we report average results for the relation and attribute categories.

\textbf{VALSE} \cite{parcalabescu2021valse} serves as a task-agnostic benchmark aimed at assessing the foundational visual-linguistic competence of general-purpose pretrained VLMs. It comprises six linguistic phenomena: existence, plurality, counting, spatial relations, actions, and entity coreference. We evaluate our method on the three subsets most relevant to visio-linguistic compositionality, in accordance with \cite{li2024interpretable}.

\textbf{What’s-Up} \cite{kamath2023s} is a spatial reasoning benchmark specifically designed to test VLMs' understanding of object spatial relation. It consists of three datasets: What’sUp ($820$ manually curated images), constructed with controlled object layouts to mitigate spatial priors. COCO-spatial ($2,687$ images), derived from the COCO dataset, pairs each image with two mutually exclusive captions differing in spatial expressions. GQA-spatial ($1,451 images$), adapted from the GQA validation set, contains spatial questions with unambiguous object references and prominent object sizes. We report the average accuracy across all three datasets.

\textbf{Hard Positive Benchmark} \cite{kamath2024hard} is introduced to measure model robustness under semantic-preserving compositional perturbations. This benchmark comprises $56,191$ images, including $28,748$ swap-based and $27,443$ replacement-based hard positives.

\subsection{Downstream Classification Benchmark}
\label{app:downstream}
Due to computational constraints, we evaluate the model's performance under both zero-shot and linear probing settings across $11$ widely used image classification datasets: CIFAR-10, CIFAR-100, Caltech-101, MNIST, VOC-2007, Aircraft, Hateful Memes, Rendered SST2, FER-2013, RESISC45, EuroSAT and FGVC-Aircraft. For linear probing, we adopt a full-shot training setup, training each model for $50$ epochs using SGD optimizer with a learning rate of $0.1$ and a weight decay of $1$e$-6$.

\subsection{Semantic Textual Similarity Benchmark}
\label{app:STS_benchmark}
STS-Benchmark \cite{cer2017semeval} is a standard dataset for semantic similarity assessment, comprising sentence pairs drawn from diverse domains such as news headlines, image captions, and QA forums. Each pair is annotated with a continuous similarity score ranging from $0$ to $5$. In contrast, SICK-R \cite{marelli2014semeval} is designed to assess compositional semantics by systematically generating sentence pairs that reflect fine-grained semantic differences induced by lexical and syntactic variations. It places a greater emphasis on a model’s ability to understand structured and compositional meaning.

\vspace{+1mm}
\section{Detailed Ablation Study of Key Designs}
\label{app:ablation}
In Table~\ref{tab:ablation} and Table~\ref{tab:strategy_ablation}, we present ablation studies analyzing key components of our framework. Due to space limitations, detailed results are provided in here. The main text discusses the impact of cross-modal masked modeling losses, auxiliary objectives,  global-to-local semantic injection, stop-gradient strategy, and the masking scheme
\begin{table*}[htbp!]
    \centering
    \resizebox{\linewidth}{!}{
        \begin{tabular}{l cccc c c c c} \toprule[1.25pt]
            Model  & $\mathcal{L}_{MLM}$  & $\mathcal{L}_{MIM}$ & $\mathcal{L}_{MAC}$ & $\mathcal{L}_{MIR}$ &  ARO  & Sugar-Crepe &  VL-Checklist  & Avg. \\ \midrule
            CLIP (ViT-B/32)  & - & - & - & -  & 58.5 & 69.8 & 65.6 & 64.6\\
            CLIP-FT& - & - & - & - & 59.9 & 74.4 & 64.2 & 66.1 \\
            \midrule
             \multicolumn{9}{c}{ablation of \texttt{cross-modal masked modeling losses}} \\
             & \checkmark &  - & - & - & 
             68.2 &	74.5 & 65.0	& 68.2 (+2.1) \\
             & - &  \checkmark &  - & - & 
             65.3 & 75.4	& 64.7	& 68.5 (+2.3) \\
            \hdashline
             & - &  - & \checkmark & \checkmark & 65.0 &	75.1 & 66.2	& 68.7 (+2.6) \\
             & \checkmark &  - & \checkmark & \checkmark & 72.0 & 77.6 & 68.7	& 72.8 (+6.7) \\
             & - &  \checkmark & \checkmark & \checkmark  & 69.9 &	74.9 &	66.8 &	70.5 (+4.4) \\
            \rowcolor{lightgray}
            \textbf{MACCO-CLIP} &  \checkmark & \checkmark & \checkmark & \checkmark & \textbf{72.5} & \textbf{78.1} & \textbf{69.5} & 
            \textbf{73.4 (+7.3)} \\
            \midrule
            \multicolumn{9}{c}{ablation of \texttt{two auxiliary losses}} \\ 
            & - &  - & - & \checkmark & 58.2 & 75.4 & 64.0 & 65.9 (-0.2) \\
            & - &  - & \checkmark & - & 63.9 & 74.4 & 65.2 & 67.9 (+1.8) \\
            \rowcolor{lightgray}
            & - &  - & \checkmark & \checkmark & 65.0 & 75.1 & 66.2 & 68.7 (+2.6) \\
            \hdashline
            & \checkmark &  \checkmark & - & - & 65.1 & 75.0 & 64.5 & 68.2 (+2.1) \\
            & \checkmark &  \checkmark & \checkmark & -  & 71.5 & 77.9 & 68.3 & 72.6 (+6.5)\\
            & \checkmark &  \checkmark & - & \checkmark  & 64.1 & 75.2 & 64.2 & 67.8 (+1.7) \\
            \rowcolor{lightgray}
            \textbf{MACCO-CLIP} &  \checkmark & \checkmark & \checkmark & \checkmark & \textbf{72.5} & \textbf{78.1} & \textbf{69.5} & 
            \textbf{73.4 (+7.3)} \\
            \bottomrule[1.25pt]
        \end{tabular}
        }
        \caption{\textbf{Ablation of different losses}. The numbers in parentheses indicate the performance gains relative to CLIP-FT.}
        \label{tab:ablation}
        \vspace{-2mm}
\end{table*}

\begin{table*}[htbp!]
    \centering
    \resizebox{\linewidth}{!}{
        \begin{tabular}{l ccc c c c c} \toprule[1.25pt]
            Model  & \makecell[c]{\texttt{Global-to-local} \\ \texttt{semantic injection}} & \makecell[c]{\texttt{Stop-grad}} & \makecell[c]{\texttt{Mask} \\ \texttt{compositional concepts}}  & ARO  & Sugar-Crepe &  VL-Checklist  & Avg. \\ \midrule
            CLIP (ViT-B/32)  & - & - & -  & 58.5 & 69.8 & 65.6 & 64.6\\
            CLIP-FT& - & - & - & 59.9 & 74.4 & 64.2 & 66.1 \\
            \midrule
            & - & \checkmark & \checkmark & 72.2 & 76.3 & 67.6 & 72.0\\
            & \checkmark & - & \checkmark & 72.4 & 76.0 & 68.0 & 72.1\\
            & \checkmark & \checkmark & - & 71.1 &  75.4 & 67.0 & 71.2 \\
            \rowcolor{lightgray}
            \textbf{MACCO-CLIP} & \checkmark & \checkmark & \checkmark & \textbf{72.5} & \textbf{78.1} & \textbf{69.5} & 
            \textbf{73.4} \\
            \bottomrule[1.25pt]
        \end{tabular}
    }
        \caption{\textbf{Ablation of global-to-local semantic injection operation, stop-gradient strategy and masking strategy}. We ablate the mask strategy with random mask, where random mask represents randomly masking image and text with a mask ratio of $75\%$ and $15\%$ following MAE \cite{he2022masked} and BERT \cite{devlin2019bert}.}
        \label{tab:strategy_ablation}
\end{table*}

Our use of the global-to-local semantic injection strategy serves two purposes: first, to provide the key and value tokens with more contextual global information; and second, to make the reconstruction learning more effective in constraining the global representation. This is particularly important because our contrastive learning objective (whether CLIP or SigLIP) mainly supervises the global representation, and most downstream tasks also rely on global representations. Therefore, even though SigLIP adopts bidirectional text attention, our strategy should still offer benefits. To further validate this assumption, we conduct additional experiments on SigLIP ViT-B/16, with results shown in Table~\ref{tab:siglip_g2l_ablation}. The results further indicate that incorporating our global-to-local semantic injection strategy improves performance in SigLIP models, although the gain is smaller than in CLIP-based models. This suggests that our strategy remains beneficial even when applied to architectures like SigLIP that use bidirectional text attention.
\begin{table*}[htbp!]
    \centering
    \resizebox{0.85\linewidth}{!}{
        \begin{tabular}{l c c c c c} 
        \toprule[1.25pt]
            Model  & \makecell[c]{\texttt{Global-to-local} \\ \texttt{semantic injection}}  & ARO  & Sugar-Crepe &  VL-Checklist  & Avg. \\ \midrule
            SigLIP (ViT-B/16) & -  & 27.4 & 62.8 & 50.9 & 47.0\\
            SigLIP-FT& - & 49.9 & 79.0 & 65.3 & 64.7 \\
            \midrule
            MACCO-SigLIP &  -  & 61.5 & 80.0 & 67.1 & 69.5 \\
            \rowcolor{lightgray}
            MACCO-SigLIP &  \checkmark & \textbf{62.5} & \textbf{80.2} & \textbf{67.5} & 
            \textbf{70.1} \\
            \bottomrule[1.25pt]
        \end{tabular}
        }
        \caption{\textbf{Ablation results of w and w/o global-to-local semantic injection on SigLIP}.}
        \label{tab:siglip_g2l_ablation}
\end{table*}

\vspace{+1mm}
\section{Further Simplified Ablation of Auxiliary Losses and Targeted Masking Strategy}
\label{app:simplified_ablation}
To more clearly isolate the contributions of the auxiliary losses and the targeted masking strategy in our method, we summarize the key ablation results in Table~\ref{tab:simplified_ablation}. In light of these results, we draw the following two conclusions:

\textbf{(1) Compositional masking outperforms random masking.} A comparison between ``Random Masking'' ($+1.4$) and ``Compositional Concept Masking'' ($+2.1$) without auxiliary losses demonstrates that specifically targeting compositional concepts is more effective than random masking. Furthermore, contrasting ``Random + Aux'' ($71.2$) with MACCO ($73.4$) highlights that while both benefit from enhanced feature representation learning, MACCO achieves an additional performance gain of $+2.2$\%. This underscores the effectiveness of our masking strategy, as compositional concepts serve as the ``structural glue'' of a scene, masking these concepts forces the model to engage in higher-order vision-language reasoning and moving beyond simple token-level reconstruction.

\textbf{(2) Compositional masking and enhanced feature representation learning work synergistically.} The performance improvement achieved by MACCO ($+7.3$) significantly exceeds the additive contributions of ``Auxiliary Losses'' ($+2.6$) and ``Compositional Masking'' ($+2.1$) individually. This substantial synergy indicates that our masked-augmented losses ($L_{MCA}$\&$L_{MIR}$) play a crucial role in effectively regularizing the feature space and facilitating the masked modeling process. This finding underscores the indispensable interplay between compositional masking and improved feature representation learning, as both components mutually reinforce each other to achieve notable performance gains.
\begin{table*}[htbp]
    \centering
    \resizebox{1.0\linewidth}{!}{
    \begin{tabular}{l ccc}
        \toprule[1.25pt]
        {Method} & 
        \makecell[c]{{$L_{MCA}$ \& $L_{MIR}$} \\ {(Improved feature} \\ {representation learning)}} & 
        \makecell[c]{{Masking} \\ {Strategy}} & 
        \makecell[c]{{Avg. Compositional} \\ {Performance}} \\
        \midrule
        CLIP & $\times$ & None & 64.6 \\
        CLIP-FT & $\times$ & None & 66.1 \\
        CLIP + Auxiliary Losses & $\checkmark$ & None & 68.7 (+2.6) \\
        CLIP + Random Masking & $\times$ & Random & 67.5 (+1.4) \\
        CLIP + Random Masking + Auxiliary Losses & $\checkmark$ & Random & 71.2 (+5.1) \\
        CLIP + Compositional Concept Masking & $\times$ & Compositional & 68.2 (+2.1) \\
        \rowcolor{lightgray}\textbf{MACCO-CLIP (ours)} & \textbf{$\checkmark$} & \textbf{Compositional} & \textbf{73.4 (+7.3)} \\
        \bottomrule[1.25pt]
    \end{tabular}
    }
    \caption{\textbf{Simplified ablation of auxiliary losses and targeted masking strategy.}}
    \label{tab:simplified_ablation}
\end{table*}
\vspace{-1mm}

\section{Freeze or Fire Image Encoder?}
\vspace{-1mm}
\label{app:ablation_not_freeze_image_encoder}
We conduct additional experiments with the vision encoder frozen, as shown in Table~\ref{tab:ablation_freeze_image_encoder}. Overall, the results indicate that finetuning the vision encoder leads to better performance, although freezing it yields marginal advantages on a few benchmarks.

This observation is reasonable regarding the findings from prior work \cite{zhai2022lit,sung2022vl,li2022blip}, which suggest that jointly optimizing both modalities enhances the flexibility of the shared embedding space and enables more effective cross-modal alignment. Specifically, \citet{zhai2022lit} suggests that while freezing the vision encoder may improve training efficiency and mitigate overfitting, it is generally more effective when the encoder is already strong, for instance, pretrained via self-supervised learning on large-scale image datasets such as JFT-300M. \citet{sung2022vl} argues that the adaptability of the vision encoder’s feature space is critical for downstream text-side tuning. If the vision encoder is fixed, its output space may be too rigid, limiting the text encoder’s ability to capture cross-modal semantics. \citet{li2022blip} shows that jointly optimizing both encoders leads to more stable and higher performance in both zero-shot and fine-tuning settings.
\begin{table}[tb]
    \centering
    \resizebox{\linewidth}{!}{
        \begin{tabular}{l c c c c c} 
        \toprule[1.25pt]
        Model  & \makecell[c]{\texttt{Fire}\\ \texttt{Image Encoder}}  & ARO  & Sugar-Crepe &  VL-Checklist  & Avg. \\ 
        \midrule
        MACCO-CLIP &  -  & \textbf{72.7} & 75.2 & 67.3 & 71.7 \\
        \rowcolor{lightgray}
        MACCO-CLIP &  \checkmark & 72.5 & \textbf{78.1} & \textbf{69.5} & \textbf{73.4}\\
        \bottomrule[1.25pt]
        \end{tabular}
        }
        \caption{\textbf{Ablation results of the choice not to freeze the image encoder.}}
        \label{tab:ablation_freeze_image_encoder}
\end{table}

\vspace{-4mm}
\section{Computational Budget}
\label{app:budget}
In Table~\ref{tab:budge}, we present the model sizes along with the training and evaluation budgets for all models discussed in our paper. Compared to standard finetuning, our method does not introduce significant additional training cost, and the inference cost remains unchanged.
\begin{table}[!h]
   \centering
   \resizebox{\linewidth}{!}{
   \begin{tabular}{l c c c}
   \toprule[1.25pt]
   Model & \#Params & Training Budge & Evaluation Budge \\
   \midrule
   \multicolumn{4}{c}{\textit{Backbone: CLIP ViT-B/32}}\\
   CLIP & 151M & -  & 0.3h \\
   CLIP-FT  & 151M & 0.8h & 0.3h \\
   SDS-CLIP & 151M & -  & - \\
   IL-CLIP  & 151M & -  & 0.3h \\
   CLIP-CAE & 151M & -  & - \\
   MACCO-CLIP & 151M & 1.0h & 0.3h \\ 
   \midrule
   \multicolumn{4}{c}{\textit{Backbone: CLIP ViT-B/16}}\\
   CLIP & 151M & -  & 0.4h \\
   CLIP-FT  & 151M & 2.5h & 0.4h \\
   MACCO-CLIP & 151M & 2.6h & 0.4h \\ 
   \midrule
   \multicolumn{4}{c}{\textit{Backbone: CLIP ViT-L/14}}\\
   CLIP & 427M & -  & 1.0h \\
   CLIP-FT  & 427M & 8.8h & 2.2h \\
   MACCO-CLIP & 427M & 9.0h & 2.2h \\ 
   \midrule
   \multicolumn{4}{c}{\textit{Backbone: SigLIP ViT-B/16}}\\
   SigLIP & 172M & -  & 1.5h \\
   SigLIP-FT  & 172M & 2.2h & 1.5h \\
   MACCO-SigLIP & 172M & 2.5h & 1.5h \\ 
   \bottomrule[1.25pt]
   \end{tabular}
}
  \caption{\textbf{Model size and computational budge.}}
  \label{tab:budge}
\end{table}

\section{Error Bar}
\label{app:error_bar}
In Table \ref{tab:error_bar}, we report the mean and standard deviation of the model performance trained using four different random seeds.
\begin{table*}[htbp]
    \centering
    \resizebox{\linewidth}{!}{
    \begin{tabular}{l ccc cc cc c c}
            \toprule[1.25pt]
            \multirow{2}{*}{Model} &\multicolumn{3}{c}{ARO} 
            &\multicolumn{2}{c}{Sugar-Crepe} 
            &\multicolumn{2}{c}{VL-Checklist} 
            &\multicolumn{1}{c}{VALSE}
            &\multicolumn{1}{c}{What's-up} 
            \\
            \cmidrule(rrr){2-4} \cmidrule(rr){5-6} \cmidrule(rr){7-8} \cmidrule(r){9-9} \cmidrule(r){10-10}
             & Relation & Attribute & Order & Relation & Attribute & Relation & Attribute & Relation & Relation   \\
            \midrule
            Random Chance &50.0 &50.0 & 20.0 & 50.0  & 50.0 & 50.0 & 50.0 & 50.0 & 41.7\\
            \midrule  
            CLIP (ViT-B/32)  & 58.7 & 62.7 & 54.1 & 68.8 & 70.8 & 63.6 & 67.7 & 70.1 & 41.8\\
            
            CLIP-FT  & 64.4$\pm$0.40 & 66.2$\pm$0.05 & 48.8$\pm$0.28 & 71.1$\pm$0.26 & 77.4$\pm$0.22 & 60.6$\pm$0.22 & 67.4$\pm$0.05 & 69.4$\pm$0.30 & 41.3$\pm$0.16 \\
            \midrule
            \rowcolor{lightgray}
            MACCO-CLIP (ours) & \textbf{73.5}$\pm$0.60 & \textbf{69.1}$\pm$0.64 & \textbf{75.0}$\pm$1.14 & \textbf{76.1}$\pm$0.95 & \textbf{78.4}$\pm$0.50 &  \textbf{69.9}$\pm$1.03 & \textbf{68.9}$\pm$0.62 & \textbf{75.2}$\pm$0.44 & \textbf{43.0}$\pm$0.72 \\
            \bottomrule[1.25pt]        
    \end{tabular}
    }
    \caption{\textbf{Multiple Runs.} We report the mean and standard deviation over four training runs of CLIP-FT and our MACCO-CLIP with four different random seeds.}
    \label{tab:error_bar}
\end{table*}

\section{More Experiments on Other Model Scales}
\label{app:other_model_scale}
We conduct experiments on three models with different scales and training paradigms: ViT-B/16, ViT-L/14, and SigLIP ViT-B/16. The results are presented in Table \ref{tab:other_model_scale}. We compare our method with CLIP-FT or SigLIP-FT for a fair comparison. 

Based on the experimental results, we have the following observations: \textbf{(1) Strong generalization:} Our method consistently improves compositional understanding across models with different scales and training paradigms (\eg, InfoNCE loss vs. pairwise sigmoid loss), demonstrating strong generalization potential. 
\textbf{(2) Greater improvements on contrastively trained models:} Compared to SigLIP ViT-B/16, CLIP ViT-B/16 exhibits larger performance gains from our method. This may be due to the fact that SigLIP does not adopt explicit batch-level contrastive learning but instead relies on pairwise contrast, while our auxiliary losses are better aligned with batch-level contrastive learning paradigms.
\begin{table*}[htbp]
    \centering
    \resizebox{\linewidth}{!}{
    \begin{tabular}{l ccc cc cc c c}
            \toprule[1.25pt]
            \multirow{2}{*}{Model} &\multicolumn{3}{c}{ARO} 
            &\multicolumn{2}{c}{Sugar-Crepe} 
            &\multicolumn{2}{c}{VL-Checklist} 
            &\multicolumn{1}{c}{VALSE}
            &\multicolumn{1}{c}{What's-up} 
            \\
            \cmidrule(rrr){2-4} \cmidrule(rr){5-6} \cmidrule(rr){7-8} \cmidrule(r){9-9} \cmidrule(r){10-10}
             & Relation & Attribute & Order & Relation & Attribute & Relation & Attribute & Relation & Relation   \\
            \midrule
            Random Chance &50.0 &50.0 & 20.0 & 50.0  & 50.0 & 50.0 & 50.0 & 50.0 & 41.7 \\
            \midrule
            \multicolumn{10}{c}{\textit{Backbone: CLIP ViT-B/16}}\\
            CLIP & 59.9 & 62.0 & 54.1 & 66.3 & 70.5 & 61.7 & 68.8 & 68.8 & 41.9\\
            CLIP-FT  & 61.2 & 62.3 & 39.9 & 71.6 & 78.5 & 56.9 & 68.5 & 67.7 & \textbf{44.2}\\
            \rowcolor{lightgray}
            \textbf{MACCO-CLIP (ours)} & \textbf{73.4} & \textbf{67.3} & \textbf{68.0} & \textbf{76.2} & \textbf{79.5} & \textbf{66.5} & \textbf{68.9} & \textbf{70.4} & {41.7}\\
            \midrule
            \multicolumn{10}{c}{\textit{Backbone: CLIP ViT-L/14}}\\
            CLIP  & 61.7 & 61.7 & 51.3 & 65.0 & 70.8 & 64.7 & 68.0 & 66.7 & 41.2 \\
            CLIP-FT  &  58.1 & 63.8 & 38.4 & 75.3 & 78.9 & 63.0 & 71.8 & 71.4 & 42.0 \\
            \rowcolor{lightgray} 
            \textbf{MACCO-CLIP (ours)} & \textbf{72.6} & \textbf{65.7} & \textbf{59.8} & \textbf{77.3} & \textbf{79.3} & \textbf{72.5} & \textbf{72.1} & \textbf{74.0} & \textbf{42.4}\\
            \midrule
            \multicolumn{10}{c}{\textit{Backbone: SigLIP ViT-B/16}}\\
            SigLIP & 26.6 & 44.6 & 11.1 & 57.9 & 67.6 & 42.0 & 59.8 & 53.5 & \textbf{41.5}  \\
            SigLIP-FT & 48.3 & \textbf{67.5} & 33.9 & 75.0 & 82.9 & 60.6 & 69.9 & 67.4 & 40.5\\
            \rowcolor{lightgray} 
            \textbf{MACCO-SigLIP (ours)} & \textbf{54.6} & {67.0} & \textbf{66.0} & \textbf{76.2} & \textbf{84.2} & \textbf{63.7} & \textbf{71.2} & \textbf{71.2} & {40.8}\\
            \bottomrule[1.25pt]
    \end{tabular}
    }
    \caption{
    \textbf{Extended experimental results on other model scales.}}
    \label{tab:other_model_scale}
\end{table*}

\section{Detailed Version of Main Experimental Results}
\label{app:subset_results}
We present the detailed results on all benchmark subsets in Table~\ref{tab:aro_detail}, Table~\ref{tab:sugarcrepe_detail}, Table~\ref{tab:vl_checklist_detail} and Table~\ref{tab:valse_and_whatsup_detail}. 
As shown, our method achieves the best performance on nearly all subsets across the five benchmarks.
\begin{table*}[htbp]
\centering
\resizebox{0.75\linewidth}{!}{
\begin{tabular}{lccccc}
        \toprule[1.25pt]
         \multirow{2}{*}{Model}& \multicolumn{5}{c}{ARO} \\
        \cmidrule(rr){2-6}
         & Relation & Attribute & COCO-Order & Flicker-Order & Avg.  \\
        \midrule
        CLIP   & 58.7     & 62.7      & 48.0       & 60.2          & 57.4 \\
        CLIP-FT  & 64.3     & 66.2      & 43.3       & 54.8          & 57.2 \\
        IL-CLIP  & 50.0     & 55.3      & 16.8       & 16.6          & 34.7 \\
        SDS-CLIP  & 53.0     & 62.0      & 24.0       & 34.0          & 43.3 \\
        CLIP-CAE  & 69.5     & 65.4      & -          & -             & -    \\
        \rowcolor{lightgray}
        \textbf{MACCO-CLIP} & \textbf{73.1}     & \textbf{68.5}      & \textbf{72.3}       & \textbf{79.6}          & \textbf{73.4} \\
        \bottomrule[1.25pt]
\end{tabular}
}
\caption{\textbf{Detailed results on ARO \cite{yuksekgonul2023when}.} In the main paper, we take the average performance of the model on the COCO-Order and Flickr-Order subsets as the performance of ARO-Order.}
\label{tab:aro_detail}
\end{table*}

\begin{table*}[htbp]
\centering
\resizebox{\linewidth}{!}{
\begin{tabular}{lcccccccccc}
    \toprule[1.25pt]
    \multirow{3}{*}{Model} & \multicolumn{10}{c}{Sugar-Crepe}  \\
    \cmidrule(rr){2-11}
    \multicolumn{1}{c}{} & \multicolumn{4}{c}{REPLACE}& \multicolumn{3}{c}{SWAP}&\multicolumn{3}{c}{ADD}\\
    \cmidrule(rr){2-5} \cmidrule(rr){6-8} \cmidrule(rr){9-11}
    & Relation& Attribte& Object& Avg.& Attribute& Object& Avg.& Attribute& Object& Avg.\\
    \midrule
    CLIP& 68.9& 80.1& 90.8& 79.9& 63.5& 60.4& 62.0& 68.4& 76.9& 72.7\\
    CLIP-FT& 71.1& 84.1& \textbf{92.9}& 82.7& \textbf{70.3}& 69.0& 69.7& 78.6& \textbf{87.2}& 82.9\\
    IL-CLIP& 56.3& 66.4& 77.7& 66.8& 54.7& 54.7& 54.7& 70.8& 65.4& 68.1\\
    \rowcolor{lightgray}
    \textbf{MACCO-CLIP}  & \textbf{77.1} & \textbf{84.6} & {92.1} & \textbf{84.6} & {70.0} & \textbf{72.2} & \textbf{71.1} & \textbf{82.8} & {86.5} & \textbf{84.7} \\
    \bottomrule[1.25pt]
\end{tabular}
}
\caption{\textbf{Detailed results on Sugar-Crepe \cite{hsieh2024sugarcrepe}.}}
\label{tab:sugarcrepe_detail}
\end{table*}

\begin{table*}[htbp]
\centering
\resizebox{\linewidth}{!}{
\begin{tabular}{lcccccccccccc}
    \toprule[1.25pt]
    \multirow{3}{*}{Model}  & \multicolumn{12}{c}{VL-Checklist}\\
    \cmidrule(rr){2-13}
    & \multicolumn{3}{c}{Relation}& \multicolumn{6}{c}{Attribute}& \multicolumn{3}{c}{Object}\\
    \cmidrule(rr){2-4} \cmidrule(rr){5-10} \cmidrule(rr){11-13}  
    & Action& Spatial& Avg.& Action& Color& Material& Size& State& Avg.& Location& Size& Avg.\\
    \midrule
    CLIP& 71.3& 55.7& 63.6& 73.3& 69.3& 66.5& 63.6& 65.4& 67.7& 77.4& 76.2& 76.8\\
    CLIP-FT& 70.7& 51.1& 60.9& 74.1& 73.1& 66.9& 60.4& 62.4& 67.4& 79.8& 78.3& 79.1\\
    IL-CLIP& 62.3& 49.0& 55.7& 64.8& 65.0& 61.9& 48.9& 57.1& 59.5& 71.1& 67.3& 69.2\\
    \rowcolor{lightgray}
    \textbf{MACCO-CLIP}  & \textbf{75.4} & \textbf{65.1} & \textbf{70.2} & \textbf{75.3} & \textbf{73.6} & \textbf{70.7} & \textbf{57.0} & \textbf{66.6} & \textbf{68.7} & \textbf{82.0} & \textbf{79.8} & \textbf{80.9} \\
    \bottomrule[1.25pt]
\end{tabular}
}
\caption{\textbf{Detailed results on VL-Checklist \cite{zhao2022vl}.}}
\label{tab:vl_checklist_detail}
\end{table*}

\begin{table*}[htbp]
\centering
\resizebox{0.85\linewidth}{!}{
\begin{tabular}{lccccccc}
    \toprule[1.25pt]
    \multirow{2}{*}{Model}  & \multicolumn{3}{c}{VALSE}& \multicolumn{4}{c}{What's-Up}\\
    \cmidrule(rr){2-4} \cmidrule(rr){5-8} 
    & Action& Relation& Avg.& Whats'Up& COCO-spatial& GQA-spatial& Avg.\\
    \midrule
    CLIP& 74.8& 65.4& 70.1& 31.1& 47.4& 46.9& 41.8\\
    CLIP-FT& 73.5& 65.1& 69.3& 30.7& 46.9& 46.5& 41.4\\
    IL-CLIP& 58.5& 52.9& 55.7& 26.0& \textbf{52.3}& \textbf{48.5}& 42.3\\
    \rowcolor{lightgray}
    \textbf{MACCO-CLIP}  & \textbf{78.6} & \textbf{72.0} & \textbf{75.3} & \textbf{34.2} & {47.1} & {48.3} & \textbf{43.2} \\
    \bottomrule[1.25pt]
\end{tabular}
}
\caption{\textbf{Detailed results on VALSE \cite{parcalabescu2021valse} and What's-up \cite{kamath2023s}.}}
\label{tab:valse_and_whatsup_detail}
\vspace{-3mm}
\end{table*}

\section{Experiments Beyond COCO Domain}
\label{app:exp_beyond_coco}
To further validate the effectiveness of our method beyond the MSCOCO dataset, we conduct experiments in two respects: on the one hand, we evaluate it on out-of-distribution benchmarks outside the MSCOCO domain; on the other hand, we train it on non-COCO datasets.

\textbf{(1) Evaluation results on Winoground and MMVP.} We conduct additional evaluations on two challenging out-of-distribution compositional reasoning benchmarks: Winoground \cite{thrush2022winoground} and MMVP \cite{tong2024eyes}. Both benchmarks consist of image-text pairs that lie beyond the COCO domain, and are widely recognized as some of the most difficult benchmarks in the field, with the results presented in Table~\ref{tab:winoground_mmvp}.
As shown, our method achieves consistent improvements over the baseline on both benchmarks. The relatively smaller gains on Winoground may be attributed to the intrinsic difficulty of the benchmark \cite{diwan2022winoground}, and evaluation on Winoground has also been noted to present out-of-distribution challenges \cite{zhang2024contrasting, li2024interpretable}. Furthermore, the performance gains on MMVP are a promising signal. Although our method primarily aims to enhance the encoding capability of the text encoder, the gains on a vision-centric task like MMVP suggest that our approach may also benefit multimodal large language models that use CLIP as the vision encoder, such as LLaVA \cite{liu2023visual}. In Section~\ref{sec:application}, we replace the vision encoder of LLaVA-1.5-7B \cite{liu2024improved} with MACCO-CLIP (ViT-L/14) and vanilla CLIP (ViT-L/14) and train under identical settings. The results in Table \ref{tab:application_mllm} show that using vision encoder from our MACCO-CLIP yields better performance in mitigating compositional semantic hallucinations.
\begin{table}[tb]
    \centering
    \resizebox{\linewidth}{!}{
    \begin{tabular}{l ccc c}
            \toprule[1.25pt]
            \multirow{2}{*}{Model} & \multicolumn{3}{c}{Winoground} & MMVP\\
            \cmidrule(rr){2-4} \cmidrule(rr){5-5}
             & Text score & Image score & Group score & Avg. \\
            \midrule
            CLIP & 31.6 & 11.1 & 9.4 & 14.8 \\
            CLIP-FT & 32.2 & 8.8 & 5.9 & 20.7 \\
            \rowcolor{lightgray}
            {MACCO-CLIP (ours)} & {32.2} & {11.1} & {8.2} & {21.5}\\
            \bottomrule[1.25pt]        
    \end{tabular}
    }
    \caption{\textbf{Experiment results on Winoground and MMVP.}}
    \label{tab:winoground_mmvp}
    \vspace{-5mm}
\end{table}

\textbf{(2) Training beyond COCO.}
To further validate the generalization capability of our method beyond the COCO domain, we conduct experiments using a subset of CC3M released by the excellent work \cite{oh2024preserving}, which contains approximately $100$k samples. We retrain both CLIP-FT and MACCO on this dataset and evaluate them on five compositional reasoning benchmarks. As shown in Table \ref{tab:trained_on_cc3m}, MACCO significantly outperforms the baseline across all five benchmarks. This further substantiates the effectiveness of our method and demonstrates that its benefits are not limited to object-centric datasets like COCO. 

This cross-domain effectiveness, combined with the OOD benchmark results, provides strong evidence that MACCO's benefits generalize beyond the specific structure of COCO.
\begin{table*}[htbp]
    \centering
    \resizebox{0.8\linewidth}{!}{
    \begin{tabular}{l ccccc}
            \toprule[1.25pt]
            \multirow{2}{*}{Model} & \multicolumn{5}{c}{Compositional Understanding Benchmarks} \\
            \cmidrule(lr){2-6}
            & ARO & SugarCrepe & VL-Checklist & VALSE & What's-up \\
            \midrule
            CLIP & 58.5 & 69.8 & 65.7 & 70.1 & 41.8 \\
            CLIP-FT & 64.5 & 75.7 & 67.5 & 70.6 & 41.2 \\
            \cellcolor{gray!20}{\textbf{MACCO-CLIP (ours)}} & \cellcolor{gray!20}\textbf{72.8} & \cellcolor{gray!20}\textbf{76.8} & \cellcolor{gray!20}\textbf{70.9} & \cellcolor{gray!20}\textbf{73.4} & \cellcolor{gray!20}\textbf{42.3} \\
            \bottomrule[1.25pt]        
    \end{tabular}
    }
    \caption{\textbf{Experimental results of models trained on CC3M.}}
    \label{tab:trained_on_cc3m}
    \vspace{-2mm}
\end{table*}

\section{Discussion with Related Works About Masked Modeling}
\label{app:masked_modeling_related_works}
MaskVLM \cite{kwon2022masked} and CLIP-CAEv2 \cite{zhang2022cae} are two studies closely related to our work. While MaskVLM is an influential work in vision-language pretraining, our method differs from it in both the masking strategy and the training objective:

\textbf{(1) Task focus and masking strategy is different.} MaskVLM focuses on multimodal pretraining and is primarily designed for general multimodal tasks, which makes the use of random masking appropriate. In contrast, our goal is to enhance fine-grained compositional understanding, which necessitates a more targeted masking strategy. Thus we introduce masking over compositional concepts spanning both text and image modalities. 

\textbf{(2) Training objective is different.} As masked signal modeling primarily constrains local tokens, whereas both contrastive learning paradigms and downstream tasks rely on global tokens, thus designing compositional masking alone is not sufficient. Therefore, we incorporate global tokens into the masked modeling process to jointly optimize global representations and facilitate reconstruction. Specifically, we introduce a global-to-local semantic injection strategy. To ensure that the masked global tokens in global-to-local semantic injection carry meaningful semantics, we further propose two masked-augmented auxiliary losses to constrain the masked global tokens. And the masking strategy of CLIP-CAE v2 is also random (applied only to the image modality).

As pretraining methods can also be adapted for fine-tuning, we conduct two additional experiments to compare our method with settings that adopt the pretraining strategies to those used in MaskVLM and CLIP-CAEv2. The results are presented in Table~\ref{tab:masked_modeling_related_works}. As shown, directly transferring the masked modeling strategies from these influential pretraining methods does not yield significant improvements, and their performance is consistently lower than ours across all benchmarks (with average gains of $1.4\%$ and $2.8\%$, compared to our $7.3\%$). These results further validate the effectiveness of our method, which uses a more targeted masking strategy, along with two auxiliary losses and a global-to-local semantic injection strategy.
\begin{table*}[tb]
    \centering
    \resizebox{0.75\linewidth}{!}{
        \begin{tabular}{l c c c c} 
        \toprule[1.25pt]
        Model   & ARO  & Sugar-Crepe &  VL-Checklist  & Avg. \\ 
        \midrule
        Baseline& 59.9& 74.4& 64.2& 66.1\\
        MaskVLM& 63.0& 74.7& 64.7& 67.5 (+1.4\%)\\
        CLIP-CAEv2& 66.0& 75.3& 65.3& 68.9 (+2.8\%) \\
        \rowcolor{lightgray}\textbf{MACCO-CLIP (ours)} & \textbf{72.5} & \textbf{78.1} & \textbf{69.5} & \textbf{73.4 (+7.3\%)} \\
        \bottomrule[1.25pt]
        \end{tabular}
        }
        \caption{\textbf{Performance comparison with models utilizing the pretraining masking strategies of MaskVLM and CLIP-CAEv2.}}
        \label{tab:masked_modeling_related_works}
\end{table*}

\section{Discussion About the Detection Model}
\label{app:robust_to_detection_model}
\begin{table*}[!h]
    \centering
    \resizebox{\linewidth}{!}{
        \begin{tabular}{l ccccccc} 
        \toprule[1.25pt]
        \multirow{2}{*}{Model} & \multirow{2}{*}{Main Detection Strength} & \multicolumn{2}{c}{ARO}       & \multicolumn{2}{c}{Sugar-Crepe} & \multicolumn{2}{c}{VL-Checklist} \\
        \cmidrule(rr){3-4} \cmidrule(rr){5-6} \cmidrule(rr){7-8}
        & & Relation      & Attribute& Relation& Attribute     & Relation& Attribute\\
        \midrule
        CLIP& -& 58.7& 62.7& 68.8& 70.8          & 63.6& 67.7\\
        CLIP-FT& \textbf{-}& 64.3& 66.2          & 71.1& 77.7& 60.9& 67.4\\
        MACCO-CLIP (OWLv2)& category-level detection& 72.0& 68.3& 76.9& \textbf{79.2} & 69.6& \textbf{69.3}  \\
        MACCO-CLIP (Grounding DINO) & \makecell[c]{language-guided visual grounding, \\category-level detection} & \textbf{73.1} & \textbf{68.5} & \textbf{77.1}  & 79.1& \textbf{70.2}& 68.7\\
        \bottomrule[1.25pt]
        \end{tabular}
        }
        \caption{\textbf{Experimental results when using an object detection model without explicit grounding training.}}
        \label{tab:detection_model_ablation}
        \vspace{-3mm}
\end{table*}
Our method primarily leverages the object detection capability of advanced grounding models to identify the object regions corresponding to grounding phrases and apply masking over the full object area. We analyze the robustness of our method for the detection model from the following two aspects.

\textbf{(1) Highly proficient visual grounding model is not a strict requirement.} Since COCO captions typically describe the prominent objects in an image, the mentioned objects are generally easy to ground. We randomly sample $100$ examples from the training set and found that only $7$ of them contained objects described in the captions that are visually ambiguous and required fine-grained attribute reasoning for accurate localization. This suggests that our method does not heavily rely on strong visual grounding capabilities. To further validate this claim, we replaced GroundingDINO with OWLv2 \cite{minderer2023scaling}, a detection model that excels at open-set object recognition without explicit grounding training. The results are show in Table~\ref{tab:detection_model_ablation}. As shown in the table, the model’s performance using masks generated by OWLv2 is comparable to that with GroundingDINO, indicating that a highly proficient off-the-shelf visual grounding model is not a strict requirement, and a general open-set detector with strong object detection capabilities is sufficient. 

\textbf{(2) Robust to noisy detection results.} To investigate the performance of our method under mild detection noise, we randomly replace the GroundingDINO's predictions with random rectangular bounding boxes with a $10\%$ probability to simulate scenarios where some of GroundingDINO's outputs might contain noise. The experimental results are shown in Table~\ref{tab:robust_to_detection_noise}. As shown in the results, even in the presence of some noise, our method still demonstrates significant improvement over the baseline, and even slightly outperforms the original data (by $+0.1\%$). This indicates that our approach exhibits a certain degree of robustness to potentially noisy grounding results, which further validates its reliability. 

This result is reasonable because when the detection model introduces noise, the resulting mask becomes random block-wise masks. Since the image and text are paired, predicting random image regions in the image based on the complete text remains plausible. Moreover, this masking strategy is more challenging than patch-wise masking, requiring the model to more deeply understand the textual information and the alignment between the image and text. Additionally, both in the fields of computer vision and natural language processing, many studies have demonstrated the advantages of block-wise masked modeling over random masking. In the field of computer vision, block-wise masked modeling has been proven to be more effective than random masking, as shown in BEiT \cite{bao2021beit}. Furthermore, in MAE \cite{he2022masked}, the authors performed ablation experiments on block-wise masked modeling (Table 1(f) in the MAE paper). The experimental results indicate that the performance of linear probing and fine-tuning with random block-wise masking pretraining is only $1.2\%$ and $1.0\%$ worse than that of random masking, respectively. This demonstrates that block-wise masking is indeed a highly effective strategy in the computer vision domain. And in the field of NLP, block-wise (or n-gram) masking is widely used in BERT-like models such as SpanBERT \cite{joshi2020spanbert} and UniLMv2 \cite{bao2020unilmv2}.
\begin{table}[b]
\centering
\resizebox{\linewidth}{!}{
    \begin{tabular}{lcccc}
    \toprule[1.25pt]
    Model& ARO  & Sugar-Crepe & VL-Checklist  & Avg.\\
    \midrule
    Baseline   & 59.9 & 74.4       & 64.2          & 66.1\\
    \makecell[l]{MACCO-CLIP \\(with 10\% noisy training data)} & 72.9 & 78.2 & 69.4 & 73.5 (+7.4\%) \\
    MACCO-CLIP & 72.5 & 78.1& 69.5 & 73.4 (+7.3\%)\\
    \bottomrule[1.25pt]
    \end{tabular}
    }
\caption{\textbf{Experimental results under noisy outputs from GroundingDINO.}}
\label{tab:robust_to_detection_noise}
\end{table}

\section{Discussion with Hard-negative Methods}
\label{app:discussion_hard_negative_method}
Addressing compositional understanding in vision-language models is a critical challenge, and would like to discuss this issue from two points:

\textbf{(1) Acknowledging contributions of prior work: } We recognize and respect the pivotal contributions made by prior works in advancing compositional understanding in VLMs. Methods such as NegCLIP \cite{yuksekgonul2023when}, TSVLC \cite{doveh2023teaching}, DAC \cite{doveh2023dense}, CE-CLIP \cite{zhang2024contrasting}, syn-CLIP \cite{cascante2023going}, IL-CLIP \cite{zheng2024iterated}, FSC-CLIP \cite{oh2024preserving},  Triplet-CLIP \cite{patel2024tripletclip}, and recent research like CLIP-CAE \cite{li2024interpretable} and SDS-CLIP \cite{basu2024distilling} have significantly progressed the field. These approaches predominantly focus on data-driven strategies, particularly through constructing hard negatives, with some methods (\eg, SDS-CLIP and CLIP-CAE) improving the loss function. These innovations have markedly enhanced vision-language compositionality and multimodal research as a whole. Additionally, benchmarks like Sugar-Crepe have been instrumental in evaluating compositional understanding, playing a crucial role in identifying and addressing the limitations of VLMs.

\textbf{(2) Our framework and its relationship to hard-negative mining:} Improving compositional understanding in VLMs remains a significant challenge, especially when relying on standard image-text pairs. While many existing methods, such as the seminal NegCLIP, focus on hard-negative mining, our approach takes a different path by emphasizing the design of improved training frameworks and loss functions. These methods should be considered orthogonal from a machine learning perspective. For instance, in alignment with SDS-CLIP and CLIP-CAE, which do not compare directly against hard-negative mining strategies, our framework integrates seamlessly with hard-negative mining methods (\eg, NegCLIP) to achieve additional gains, as demonstrated in Section \ref{sec:combined_with_hard_negative} of the main text. This compatibility further validates the effectiveness of our design and highlights the complementary nature of these approaches.

\section{Discussion About Efficacy on Spatial Relationships}
\label{app:discussion_spatial_relation}
For spatial relationships, on the text side, we can accurately mask the spatial relationship expressions, which enables the model to learn to better understand spatial relations in the image through cross-modal masked modeling. On the image side, while spatial relations cannot be directly grounded to specific regions, we can mask the regions corresponding to the two related objects and use the full text to guide reconstruction. Our motivation is that if the model understands the spatial relationship described in the text (\eg, ``object A is to the left of object B''), it should be able to reconstruct the relative positions of the two objects. In this way, we aim to enhance the model’s ability to interpret spatial relationships described in text. We present the performance of our method on several benchmark subsets specifically designed to evaluate spatial relation understanding in the Table~\ref{tab:spatial_relation_performance}. As shown, our method brings significant improvements in this type, indicating its effectiveness in enhancing the model’s comprehension of spatial relationships.
\begin{table*}[!h]
\centering
\resizebox{0.85\linewidth}{!}{
    \begin{tabular}{lccc}
    \toprule[1.25pt]
    Model & \makecell[c]{VL-Checklist \\(spatial relation subset)} & \makecell[c]{VALSE \\ (spatial relation subset)} & \makecell[c]{What’s-up \\(designed for evaluate spatial relation)} \\
    \midrule
    CLIP & 55.7 & 65.4 & 41.8 \\
    CLIP-FT & 51.1 & 65.1 & 41.4 \\
    \midrule
    \rowcolor{lightgray}MACCO-CLIP & \textbf{65.1 (+14)} & \textbf{72.0 (+6.9)} & \textbf{43.2 (+1.8)} \\
    \bottomrule[1.25pt]
    \end{tabular}
    }
\caption{\textbf{Performance on several benchmark subsets specifically designed to evaluate spatial relation understanding.}}
\label{tab:spatial_relation_performance}
\vspace{-3mm}
\end{table*}

\section{Discussion on Language Generalizability and Tool Dependency}
\label{app:discussion_tool_dependency}
Our framework leverages external tools for concept extraction, which may raise question about their availability and accuracy across diverse languages. We discuss the generalizability and robustness of our approach from two key perspectives:

\textbf{(1) Modular flexibility for multilingual support.} While our current implementation leverages a specific English scene graph parser \cite{wu2019unified}, the MACCO framework is inherently modular and not restricted to any particular legacy tool. For non-English or low-resource languages, compositional concepts (objects, attributes, and relations) can be effectively extracted using modern open-source large language models (\ie, Qwen) or multilingual NLP toolkits (\ie, Stanza or spaCy). Recent studies \cite{do2024zela, do2025enhancing} demonstrate that LLMs are highly proficient in zero-shot compositional parsing across diverse languages, underscoring the broad applicability of our approach to multilingual scenarios.

\textbf{(2) Resilience to imperfect tools.} From a visual perspective, our in-depth analysis in Appendix \ref{app:robust_to_detection_model} demonstrates that MACCO achieves robust performance even when the detection model is imperfect. From a textual perspective, if the parser fails to accurately identify compositional concepts in certain languages, our masking strategy degrades to a random masking approach at worst. As revealed in our ablation study (Section \ref{sec:ablation}), even under random masking, our framework consistently outperforms the baseline, although masking compositional concepts yields the most substantial improvement. This empirical evidence highlights that MACCO is resilient to parsing inaccuracies, leveraging these tools as a guided prior that remains effective even in the presence of partial noise.

\section{Discussion of Potential Biases from Pre-trained Tools}
\label{app:discussion_tool_bias}
MACCO leverages external tools to perform compositional concept extraction. Although these tools are well-established in practice, they inevitably introduce certain systemic biases such as neglecting long tail patterns. We discuss how MACCO mitigates the potential impact resulting from these limitations from the following perspectives.

\textbf{(1) Implicit constraints on rare compositional patterns:} MACCO does not solely rely on the ``labels'' provided by parsers, it uses these labels to generate masking signals that guide cross-modal reconstruction. Even if pre-trained tools overlook rare compositional patterns, the global features from the full and masked signals still retain information about these patterns. Our proposed global-to-local semantic injection operation integrates global semantic features into local tokens during reconstruction. This mechanism introduces implicit constraints on rare compositional patterns, thereby mitigating the impact of parser limitations.

\textbf{(2) Feature space regularization facilitating rare compositional pattern learning:} The two masked-augmented losses we propose ($L_{MCA}$\&$L_{MIR}$) further regularize the feature space, acting as an additional implicit constraint. This ensures that the model's embedding space is grounded in the entire data distribution of natural image-text pairs, rather than being overfitted to the frequent patterns disproportionately emphasized by pre-trained tools. As a result, our approach prevents the model from losing sensitivity to infrequent but meaningful compositional patterns.

\section{Discussion with X-VLM}
\label{app:discussion_with_xvlm}
X-VLM \cite{zeng2022multi} is a pioneering work in multi-grained vision-language pre-training. In contrast, MACCO introduces a fundamental paradigm shift from ``Explicit Alignment via Localization'' to ``Implicit Alignment via Reconstruction''.

\textbf{Visual Concept Localization vs. Compositional Concept Reconstruction:} X-VLM relies on the model’s ability to ``point'' to where a concept resides in the image using explicit bounding boxes. In contrast, MACCO emphasizes compositional reasoning. By strategically masking compositional anchors instead of random tokens, we encourage the model to infer missing structural dependencies through cross-modal context. This approach is inherently more demanding than localization, as it necessitates that the model internalizes how attributes and relations connect and bind to objects.

\textbf{General MLM in Language vs. Targeted Masking in Vision and Language:} While X-VLM includes a general MLM loss, MACCO adopts a more targeted approach by extracting and masking compositional concepts. Our approach addresses the ``bag-of-words'' bias in contrastive models by ensuring that the model cannot solve the reconstruction task without understanding the contextual interplay between concepts across modalities.

\textbf{Future Paths for Enhancing MACCO with Multi-Grained Reasoning:} Beyond these distinctions, we believe that X-VLM's multi-grained framework and MACCO’s implicit reconstruction paradigm are highly complementary. For instance, X-VLM’s multi-grained reasoning could naturally integrate with MACCO by introducing multi-grained reconstruction strategies, such as reconstructing compositional text while considering the global image or corresponding image regions. Furthermore, X-VLM’s multi-grained contrastive loss could be adapted to anchor compositional concepts (\eg, specific attribute-object pairs) to their associated visual regions.

\section{Discussion About the Attribute Binding Challenge}
\label{app:attribute_binding_discussion}
Attribute binding is indeed a persistent bottleneck in the field of multimodal learning. We would like to discuss this from two perspectives:

\textbf{(1) Why attribute binding is fundamentally challenging.} Attribute binding is inherently more difficult than relation modeling because attributes (\eg, color, material, size) are often visually and semantically entangled with the objects they modify. In contrastive VLMs, models can exploit shortcuts and exhibit a ``bag-of-words'' behavior, detecting the presence of individual concepts (\eg, ``red'', ``car'') without encoding the structural association that binds the attribute (\eg, ``red'') to the correct object (\eg, ``car''). Consistent with findings in compositional text-to-image synthesis \cite{huang2025t2i,chefer2023attend,rassin2023linguistic} and VLM research \cite{johnson2017clevr}, attribute binding remains a persistent challenge. Attribute cues are often spatially fused with object evidence in the same image regions, which makes them difficult to disentangle. By contrast, relations (\eg, ``on'', ``next to'') typically have clearer geometric signatures.

\textbf{(2) Preliminary ideas for future enhancements on attribute binding.} 
First, \textit{optimizing attention mechanisms.} During training, selectively blocking attention interactions across different attribute-phrase groups can reduce feature entanglement among attributes. 
Second, \textit{token merging for attribute binding.} During inference, merging tokens that correspond to the same attribute phrase in the VLM encoder can encourage stronger binding between attributes and their associated objects. 
Third, \textit{synthetic data generation.} Leveraging powerful LLMs and text-to-image generation models to synthesize datasets enriched with diverse attribute combinations can improve generalization to rare attributes. 
Fourth, \textit{enhanced text encoders.} Employing medium-sized LLMs as the text encoder can strengthen the model's ability to parse and understand complex attribute bindings in text.

\section{Discussion About Future Work Using Strong LLM-based Text Encoders}
\label{app:discussion_llm_based_text_encoder}
 \citet{zhang2025assessing} highlights that the text encoder trained under the original CLIP paradigm has limited language understanding capabilities. In contrast, incorporating more powerful pretrained language models as the text encoder presents a promising direction for building robust foundation vision-language models. We discuss this from the following two perspectives:

\textbf{Advantages of using LLM as text encoder:}
\textbf{(1) Stronger language understanding and reasoning capabilities.} LLMs possess richer syntactic, semantic, and contextual modeling abilities, enabling them to capture complex linguistic structures. This can significantly enhance a model's understanding of compositional relations and semantic nuances, which is particularly important for tasks that require complex reasoning. \textbf{(2) Better generalization ability with longer context.} LLMs are typically trained on large-scale, open-domain corpora, making them more robust to rare vocabulary, long-tail compositional patterns, and and more effective to understand long sentences containing complex linguistic structures. As a result, they tend to generalize better in zero-shot and open-world settings.

\textbf{Challenges of using LLM as text encoder:}
\textbf{(1) Higher computational cost and deployment complexity.} Compared to CLIP's original lightweight text encoder, using an LLM significantly increases the number of parameters, training cost and inference latency, potentially limiting deployment in real-time scenarios. Efficient training strategy like SAIL proposed in the paper provides a promising solution to mitigate this issue. \textbf{(2) Potentially increased difficulty in cross-modal alignment.} Although LLMs produce powerful semantic representations, these may not align naturally with visual feature spaces. Especially without joint training or fine-tuning, there may be large semantic gap between modalities, which can degrade image-text alignment and contrastive learning efficiency.

In summary, replacing the CLIP-style text encoder with a strong LLM holds significant potential for improving language understanding and overall generalization in vision-language models, especially for tasks requiring complex compositional reasoning. It is a promising and increasingly recognized direction. Nonetheless, this approach also introduces new challenges related to cross-modal alignment and resource demands. In the future, we also hope to extend our framework to LLM-based CLIP-like models to further explore this direction.
\end{document}